\documentclass{apple/applemlr}

\usepackage{amsmath}
\usepackage{enumerate}
\usepackage{algorithm}
\usepackage{algpseudocode}
\usepackage{amsfonts}
\usepackage{amsthm}
\usepackage{cleveref}
\usepackage{diagbox}
\usepackage{colortbl}
\usepackage{amssymb}
\usepackage{xspace}
\usepackage{wrapfig}
\usepackage{adjustbox}
\usepackage{tabularx}
\usepackage{booktabs}
\usepackage{mathtools}
\usepackage{tikz}
\usepackage{enumitem}
\usepackage{silence}
\usepackage{dsfont}
\usepackage[table]{xcolor}
\usepackage[dvipsnames]{xcolor}
\usepackage{multirow}
\usepackage{makecell}
\usepackage{xfakebold}
\usepackage{amsmath,amsfonts,bm}

\def\eqref#1{equation~\ref{#1}}

\def\1{\bm{1}}

\def\vs{{\bm{s}}}

\DeclareMathAlphabet{\mathsfit}{\encodingdefault}{\sfdefault}{m}{sl}
\SetMathAlphabet{\mathsfit}{bold}{\encodingdefault}{\sfdefault}{bx}{n}

\newcommand{\R}{\mathbb{R}}

\definecolor{textgray}{HTML}{6E6E73}
\usetikzlibrary{positioning, calc}
\usetikzlibrary{decorations.pathmorphing}

\makeatletter
\patchcmd{\wrong@fontshape}{\@gobbletwo}{}{}{}
\makeatother
\numberwithin{equation}{section}
\makeatletter
\AtBeginDocument{
  \urlstyle{sf}
  
}
\makeatother

\definecolor{light}{RGB}{125, 125, 125}
\crefname{tcb@cnt@pbox}{code}{code}
\Crefname{tcb@cnt@pbox}{Code}{Code}
\crefname{assumption}{assumption}{assumption}
\Crefname{assumption}{Assumption}{Assumptions}

\newtcolorbox[auto counter]{pbox}[2][]{
  colback=white,
  title=Code~\thetcbcounter: #2,
  #1,fonttitle=\sffamily,
  fontupper=\sffamily,
  arc=2pt,
  colframe=bgcolor,
  coltitle=fgcolor,
  colbacktitle=bgcolor,
  toptitle=0.25cm,
  bottomtitle=0.125cm
}

\makeatletter
\newcommand\applefootnote[1]{%
  \begingroup
  \renewcommand\thefootnote{}%
  \renewcommand\@makefntext[1]{\noindent##1}%
  \footnote{#1}%
  \addtocounter{footnote}{-1}%
  \endgroup
}
\makeatother

\definecolor{cverbbg}{gray}{0.90}

\usepackage[utf8]{inputenc}
\usepackage[T1]{fontenc}
\usepackage{hyperref}
\usepackage{url}
\usepackage{booktabs}
\usepackage{amsfonts}
\usepackage{nicefrac}
\usepackage{microtype}
\usepackage{xcolor}
\hypersetup{
    colorlinks=true,
    allcolors=blue!70!black,
}
\usepackage{amsmath}
\usepackage{xspace}
\usepackage{enumitem}
\usepackage{graphicx}
\usepackage{xcolor,colortbl}
\usepackage{multirow, multicol}
\usepackage{adjustbox}
\usepackage{subcaption}
\usepackage{cleveref}

\newcommand{\pcd}{\mathcal{X}}

\newcommand{\transform}{\mathcal{M}}

\newcommand{\voxel}{\mathcal{V}}

\newcommand{\decoder}{\mathcal{D}}
\newcommand{\encoder}{\mathcal{E}}

\def\1n{\mathbf{1}_n}
\def\0{\mathbf{0}}
\def\1{\mathbf{1}}

\def\R{{\mathbb R}}

\def\f{{\mathbf{f}}}

\def\s{{\bf s}}

\def\x{{\bf x}}
\def\y{{\bf y}}

\makeatletter
\DeclareRobustCommand\onedot{\futurelet\@let@token\@onedot}
\def\@onedot{\ifx\@let@token.\else.\null\fi\xspace}

\def\eg{\emph{e.g}\onedot} 
\def\ie{\emph{i.e}\onedot} 
 
 \def\vs{\emph{vs}\onedot}

\makeatother

\newcommand{\first}{\cellcolor{red!25}}
\newcommand{\second}{\cellcolor{orange!25}}

\let\cite\citep

\title{Split-and-scale Latent 3D Representations}

\author{Dorian Chan}
\author{Muhammed Kocabas}
\author{Xiaoming Zhao}
\author{Oncel Tuzel} 
\author{Jen-Hao Rick Chang}

\affiliation{Apple}

\abstract{Latent 3D representations have significantly expanded the capabilities of modern shape modeling, enabling compact encoding of 3D scenes and supporting powerful 3D generative models.
Despite this progress, current representations struggle to capture the fine details of complex objects, fundamentally limited by the capacity of their latents.
However, scaling latent capacity at training time is computationally expensive and often infeasible under memory limits, and naively increasing the number of latents at inference fails to make effective use of the added capacity.
In this paper, we address both challenges with \textit{split-and-scale}, a method that effectively expands latent capacity at inference while keeping training cost tractable.
Our key observation is that by representing 3D scenes as continuous functions of 3D coordinates, a single pretrained tokenizer can be applied at inference to arbitrarily small sub-regions of a scene, allocating the same latent capacity to a smaller spatial extent and thus capturing far higher detail.
We validate this insight by conducting extensive ablations to study the trade-offs between compute, memory, latent capacity, and reconstruction quality, demonstrating our inference-time scaling method achieves quality competitive with much larger models that exceed single-GPU memory limits.
We further demonstrate that split-and-scale supports high-quality generative modeling, training an autoregressive 3D generator that performs competitively with state-of-the-art baselines.}

\begin{document}

\maketitle

\begin{figure}[h]
	\centering
	\includegraphics[width=\linewidth]{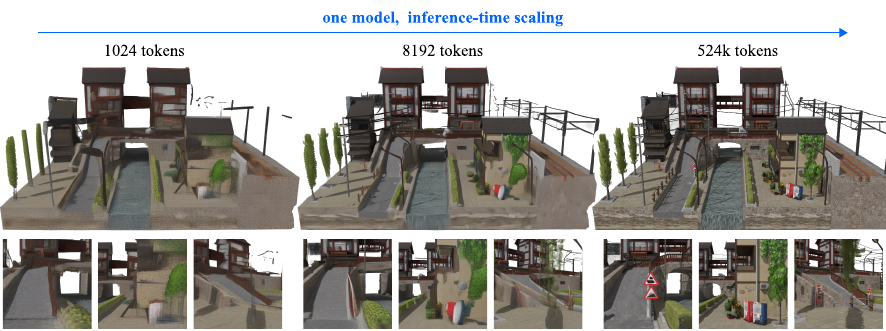}
	\vspace{-6mm}
	\caption{
		\textbf{Split-and-scale} reuses the \textit{same} tokenizer across spatial scales to effectively increase latent capacity at \textit{inference} time.  Recommend viewing the figure on monitor, zoomed in.  Asset credit~\cite{tokyo_street}.
	}
	\label{figure:teaser}
\end{figure}
\section{Introduction}
\label{sec:intro}

Latent representations have become central to modern 3D shape modeling~\cite{Zhang20233DShape2VecSetA3, chang20243d, xiang2025trellis}.
By encoding geometry and appearance into a compact set of tokens, they support a wide range of tasks, from shape reconstruction to 3D generation~\cite{zhao2023michelangelo, lito, xiang2025trellis2, zhao2025hunyuan3d, chen2025ultra3d, li2025sparc3d, lai2025lattice, jia2025ultrashape, chen2025sam3d, wu2025direct3d,malik2026velox}.

Capturing the fine details of real-world objects, however, demands far more representational capacity than current latent representations typically provide. For example, consider building a latent representation for a small town, with buildings and trees and signs (\Cref{figure:teaser}). Clearly, very large encodings are likely necessary to accurately capture all the detail in such a scene.
How can we scale the latent representations of today to handle such a scenario? Since capacity in these token-based architectures is governed by the number of latent tokens, the standard recipe is to scale up the token count.
This recipe is far from free: computation grows quadratically and memory linearly with the number of tokens, and existing mitigations like curriculum learning~\cite{wu2025direct3d} or sophisticated attention patterns~\cite{lito} only partially help --- GPU memory ultimately caps the trainable latent size (\eg, $\sim$8192 tokens for LiTo on an 80GB H100).
While certain architectures (\eg, Perceiver-style encoders~\cite{Zhang20233DShape2VecSetA3,lito,malik2026velox,lai2025lattice}) permit larger token counts at inference than at training, in practice the resulting quality gains saturate quickly and can even reverse when scaled too far (\Cref{figure:tokenizer_curves}).

In this paper, we propose \textit{split-and-scale} (\Cref{figure:teaser}), a technique that enables effective inference-time scaling of latent capacity.
Our key observation is that a latent 3D representation parameterized by continuous coordinates can be applied at any spatial scale: the same encoder operates on a full scene or any sub-region, after appropriate normalization of the input coordinates.
Critically, normalizing a sub-region of width $1/n$ back to $1$ lowers the content frequency seen by the encoder by a factor of $n$.
The same network thus represents detail at $n\times$ finer spatial resolution within each sub-region, and produces proportionally more total tokens across the full scene --- effectively increasing latent capacity with no change to the network or its weights.
In practice, because 3D scenes are dominated by surfaces, only $O(n^2)$ of the $n^3$ sub-regions are non-empty --- so the total token count grows with surface area rather than volume, keeping inference cost tractable even at fine subdivisions.

Beyond inference, split-and-scale also enables efficient training.
Because the encoder operates on normalized sub-regions, it can be trained primarily at coarser subdivisions and still generalize to finer ones at inference, as we will demonstrate empirically.
Furthermore, we encode and decode each sub-region independently while computing the reconstruction loss on the assembled full scene.
Independent encoding lets us backpropagate gradients through only a small subset of sub-regions per iteration, sharply reducing compute and memory; the full-scene loss simultaneously teaches the network to handle inter-region boundaries.
Together, these properties let split-and-scale train comfortably within a single 80GB GPU (\eg, A100 and H100), even at token counts that would otherwise be infeasible to train under standard architectures.

Through extensive reconstruction experiments, we show that split-and-scale consistently outperforms naive inference-time scaling and approaches the quality of much larger baselines that cannot be trained within a single 80GB GPU.
Detailed ablations further analyze the design choices behind split-and-scale and characterize the trade-offs between latent capacity, training compute, memory, runtime, and reconstruction quality.
We additionally demonstrate the representation's broader applicability by training an autoregressive 3D generative model on top of split-and-scale latents.

In summary, our main contributions are:
\begin{itemize}[leftmargin=*]
    \item We propose \textit{split-and-scale}, a latent 3D representation that reuses a single tokenizer across spatial scales --- enabling memory- and compute-efficient training while supporting much higher-detail reconstruction at inference.
    \item Through thorough ablation studies, we map the trade-offs between latent capacity, compute, memory, and reconstruction quality, and show that split-and-scale consistently outperforms naive inference-time scaling of the underlying tokenizer.
	 \item We show a potential use case of the representation by training an autoregressive 3D generative model on top of split-and-scale latents.
\end{itemize}

\section{Related work}

In recent years, a significant body of work has started to explore the landscape of 3D latent representations. For instance, in the context of geometry, a number of approaches explore the encoding of signed-distance functions or occupancy fields~\cite{Zhang20233DShape2VecSetA3,zhang2024clay,li2025triposg,he2025triposf,zhao2025hunyuan3d,meng2025lt3sd,hui2024make}, often over some form of sparse grid structure for efficiency~\cite{wu2024direct3d,wu2025direct3d,ren2024xcube,li2025sparc3d,lai2025lattice}. Other works attempt to model 3D surfaces as probability distributions, leveraging generative architectures to sample from them~\cite{cai2020learning,chang20243d,yang2019pointflow}. 

In recent work, latent representations for both geometry and appearance have seen more traction. A number of approaches have explored how to encode both meshes and their textures. For example, many geometry-only representations have been extended with multi-view texture generation pipelines to add material to generated assets~\cite{zhang2024clay,zhao2025hunyuan3d,li2025triposg,yang2025pandora3d}; however, such approaches do not leverage any potential compression in terms of shared information between geometry and appearance. To tackle this aspect, recent work has explored more joint encodings by baking the geometry and material of an input mesh into an intermediate primitive before latent compression. 3DTopia-XL~\cite{chen20253dtopia} proposed a PrimX primitive that locally encodes both SDF and material parameters, on top of which they train a variational autoencoder (VAE)~\cite{kingma2013auto} for further compression. TRELLIS-2~\cite{xiang2025trellis2} proposed a dual contouring-like O-Voxel representation that they then train a VAE upon. While these mesh-based approaches often produce high-quality, plausible assets, they typically fail to reconstruct the appearance of input objects even under ground truth lighting, as they rely on simplified material models that do not capture the complexities of not only real objects but also many synthetic assets. Furthermore, generative models trained upon these representations do not output the surrounding lighting. This makes such models impractical for tasks like digital tourism or digital memory preservation, where a user would need to guess the lighting that best reproduces their target scene.

Concurrently, another line of research has explored directly encoding multi-view RGBD images as a more accessible proxy of both geometry and appearance. TRELLIS~\cite{xiang2025trellis} proposed encoding voxelized DINOv2 features~\cite{oquab2023dinov2} into a structured latent representation, where features are encoded over a sparse grid --- the representation is then supervised via renderings of decoded 3D Gaussians. SAM 3D~\cite{chen2025sam3d} built a generation pipeline for a TRELLIS-style representation with a sophisticated post-training process in order to maximize generation quality, improve input image fidelity and handle occlusions.
To improve visual realism, LiTo~\cite{lito} proposed using the surface light field as the input encoder feature, and decoding to view-dependent Gaussians. Relative to TRELLIS, this allowed their method to produce much better highlights and reflections, and also enabled the training of a generative model with much stronger input image fidelity thanks to the continuous nature of the surface light field input. Velox~\cite{malik2026velox} extended the LiTo architecture to 4D scenes, demonstrating strong performance in tracking and generation tasks. 
Broadly speaking, our work is mostly orthogonal to these directions for improving latent 3D representations, and can potentially be used to scale any of them in a computationally efficient manner. In this paper, we chose to build off LiTo~\cite{lito} as a base thanks to its strong performance in reconstruction and generation tasks --- we discuss more details of how we leverage it in the next section.

More related to our work, a few approaches have begun exploring how objects can be generated in a part-based fashion~\cite{tang2026efficient,chen2025ultra3d,lin2026partcrafter,he2026unipart}, typically by using different sets of latent tokens for different components. However, these approaches do not consider scaling the overall resolution of the entire object like we do, and require part-based annotations that may limit available 3D training data. Perhaps most similar to our approach, a few world-building approaches have adapted object-level tokenizers and generators, e.g., TRELLIS~\cite{xiang2025trellis}, by applying them at the block level and combining the results to compose entire scenes~\cite{engstler2025syncity,zheng2025constructing,yoon2026extend3d,li2026worldgrow}. While analogous in principle, unlike our work, they are focused on scene generation, and must handle seams between different blocks in ad hoc fashions (\eg, via RePaint~\cite{lugmayr2022repaint} or SDEdit~\cite{meng2021sdedit}) because they reuse pre-existing object tokenizers. In contrast, our work focuses on scaling the quality of object representations by building a tiling-friendly tokenizer and generative model from ground up. Ultimately, we believe our pipelines and trained models may be useful for future such world-building approaches.

\section{Method}
\label{sec: method}

\begin{figure}[t]
    \centering
    \includegraphics[width=\linewidth]{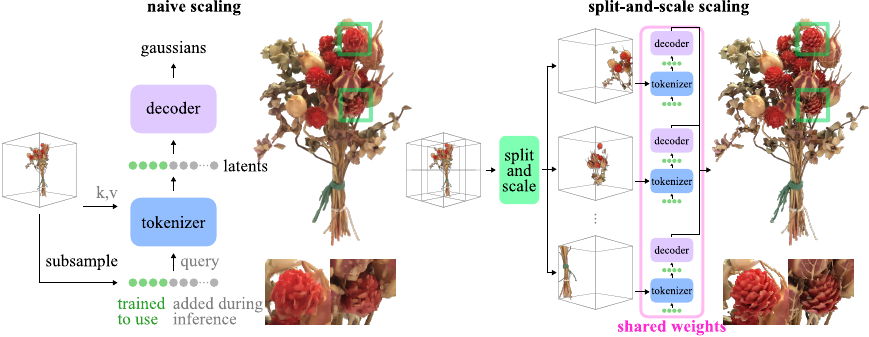}
    \caption{\textbf{Naive scaling \vs split-and-scale.} (Left) Naive scaling increases number of latents by using more queries to the tokenizer. As the tokenizer is not trained with the higher number of queries, the quality improvement is limited. (Right) Split-and-scale encodes/decodes subregions of the scene with the same tokenizer/decoder and the designed number of queries, effectively increasing latent capacity. Asset credit~\cite{dryflower}.}
    \label{figure:method}
\end{figure}

\subsection{Preliminary: LiTo~\cite{lito}}

While split-and-scale can potentially be applied to various existing methods, we focus on LiTo~\cite{lito}, a recent 3D latent representation for geometry and (view-dependent) appearance, as it takes a point cloud with continuous coordinates as input.
Given a set of points $\pcd = \{(\x_i, \f_i)\}_{i=1 \dots n}$, where $\x_i \in \R^3$ is the 3D coordinate and $\f_i \in \R^c$ is the associated feature containing color and view direction, a Perceiver encoder~\cite{jaegle2021perceiver, Zhang20233DShape2VecSetA3} $\encoder$ takes $\pcd$ as keys and values and subsamples $\pcd$ as the initial queries. It then outputs the latent representation $\s$.
The latent representation $\s$ is composed of a set of tokens, $\{s_i \in \R^d \}_{i=1 \dots m}$, and can be decoded by different decoders into point clouds and Gaussian splats.
At a high level, we have:
\begin{equation}
    \{(\x_i, f_i) \}_i \ \overset{\encoder}{\longrightarrow} \ \s \ \overset{\decoder}{\longrightarrow} \ \y,
\end{equation}
where $\y$ can be a point cloud, a mesh, or Gaussian splats. 
$\s$ is supervised via 1) novel view renderings from the decoded Gaussians and 2) points sampled from a flow-matching decoder~\cite{chang20243d}, given a dataset of RGBD images rendered and points sampled from a collection of synthetic objects. 
Mesh and voxel occupancy decoders are later trained with a frozen encoder, similar to TRELLIS~\cite{xiang2025trellis}.

\begin{figure}[t]
    \centering
    \includegraphics[width=\linewidth]{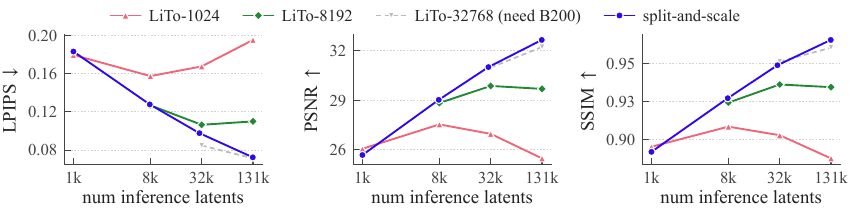} 
    \caption{\textbf{Quantitative comparison versus naive token count scaling.} We train a split-and-scale model that outputs 1024 latents per voxel. 
    We compare with LiTo-1024, LiTo-8192, LiTo-32768 that use 1024, 8192, 32768 tokens during training, respectively. We also evaluate these models with naive inference-time scaling. 
    Note that training LiTo-32768 requires larger memory than all other methods, and its training takes $2.7\times$ more time.
    All models are trained to the same number of iterations. As shown, the performance of the proposed method (blue curve) increases consistently as we increase number of tokens during inference, whereas naive scaling quickly hits bottleneck.
    }
    \label{figure:tokenizer_curves}
\end{figure}

To scale LiTo to represent more complex objects, one potential solution is to simply increase the number of latent tokens $m$ (by increasing the number of queries subsampled from $\pcd$).
However, it is often infeasible during \textit{training}, as the compute grows quadratically due to the use of attention and memory linearly (see \Cref{table:tokenizer_compute}). 
Thereby, existing works~\cite{zhang2024clay, li2025triposg} often increase number of latents during \textit{inference} by sampling more queries from the input point cloud. 
However, because the tokenizer is not trained with additional number of queries, the improvement is limited and can negatively impact the quality if used aggressively, as shown in \Cref{figure:tokenizer_curves}.

\subsection{Split-and-scale 3D Representations}

\label{subsec:splitscale}

\begin{table}[t]
    \centering
    \def\maskedcolor{gray}
    \caption{\textbf{Training compute comparison versus naive token count scaling.} Our split-and-scale recipe induces a similar computational and memory cost to an 8192 token version of LiTo~\cite{lito}, and can be comfortably trained on an 80GB GPU unlike a 32768 token LiTo. Runtime is computed over a single forward + backward pass. FLOPs are computed assuming $2\times$ forward cost for backward.}
    \label{table:tokenizer_compute}
    \adjustbox{max width=0.9\columnwidth}{%
    \begin{tabular}{ccccc}
       \toprule
       \textbf{Method} & \textbf{Peak memory (GB)} & \textbf{Avg runtime (s)} & \textbf{Avg GFLOPs} & \textbf{LPIPS (32768 inf. tokens)}\\
       \midrule
       LiTo (8192 tokens) & 68.22 & 1.18 & 45.51 & 0.106\\

       \textit{Split-and-scale} ($\approx$ 14k train. tokens) & 73.51 & 1.32 & 49.71 & 0.091\\

       \textcolor{\maskedcolor}{LiTo (32768 tokens)} & \textcolor{\maskedcolor}{145.58} & \textcolor{\maskedcolor}{3.54} & \textcolor{\maskedcolor}{306.23} & \textcolor{\maskedcolor}{0.084}\\
        \bottomrule
    \end{tabular}%
    }
\end{table}

Can we effectively increase the number of latents during inference while maintaining training efficiency?
Our strategy is to keep encoder operating exactly as at training --- with the same number of queries $m$ --- and invoke the tokenizer multiple times, once per sub-region of the scene (see \Cref{figure:method}). 
Concretely, we subdivide the scene (ranging from -1 to 1) into $n^3$ non-overlapping voxels of size $\frac{2}{n}$ and encode each independently:
$
    s_\text{split-scale}[i] = \encoder(\voxel_i),
$
where $\voxel_i$ is the content of the $i$th voxel. 
To decode, we similarly run the decoder independently for each voxel, and then compute losses over the re-assembled full scene.
For efficiency, we skip empty voxels, so the total number of latents grows directly with the number of occupied voxels --- due to the surface-dominated nature of 3D scenes, only $\mathcal{O}(n^2)$ of the $n^3$ voxels are occupied in practice. 
Output resolution (\eg, number of Gaussians produced by the decoder) also grows correspondingly with $n$.

However, splitting alone is not enough: as reported in \Cref{table:ablation_recon} (``no scaling''), LPIPS degrades sharply from $0.107$ to $0.161$ at $n=4$. 
The cause is a frequency mismatch. 
A voxel of width $\frac{1}{n}$ packs the same amount of geometric and textural detail into a spatial extent $n$ times smaller, so the encoder is fed content at substantially higher spatial frequencies than any it was exposed to during training. 
As $n$ grows, the input distribution drifts out-of-domain and quality collapses.

We correct this mismatch via the space-frequency duality of the Fourier transform: dilating a signal in space by a factor of $n$ compresses its spectrum by the same factor. 
We therefore shift and rescale each voxel back to the original scene extent $[-1, 1]^3$ with a normalization $\transform_i$ before encoding, and invert it at the decoder:
\begin{equation}
    s_\text{split-scale}[i] = \encoder(\transform_i(\voxel_i)), \qquad
    \decoder_\text{split-scale}(s_\text{split-scale}) = \bigcup_i \transform^{-1}_i\!\left(\decoder(s_\text{split-scale}[i])\right).
\end{equation}
Note that the transformation $\transform_i$ only depends on the size and location of the voxel, not its content. 
Under this rescaling, every voxel presents the encoder with content whose frequency range matches what it saw at training, regardless of voxel size. 

Interestingly, the same network captures $n\times$ finer detail inside each voxel while reusing its pretrained weights, effectively expanding total latent capacity at inference without any retraining.
By choosing $n$, we directly control the level of detail.

\subsubsection{Computational strengths of split-and-scale}
\label{sec:cstrength_sands}

Strictly speaking, a traditional train-time-scaling approach for $n$ tokens (\ie, directly training a tokenizer to use an increased number of tokens) should outperform a split-and-scale representation at $n$ tokens. Under split-and-scale, tokens are partitioned across subregions and restricted to intra-region attention, while tokens in a traditional approach can attend globally across the entire object allowing for more potential compression.
However, when implemented appropriately, the proposed method has important computational strengths, as we will discuss below:

\paragraph{Coarse-scale training.}\noindent As the encoder and decoder functions are fixed across all scales, they can be trained at coarser scales only, but still generalize to finer subdivisions at inference. 
The smaller number of latent tokens at coarser scales allows for much faster training and reduced memory usage.
In practice, slight distributional differences can be introduced due to subdividing the scene. 
First, at $n > 1$, geometry can cross different voxels, and the latent encoder/decoder must therefore be able to handle inter-voxel boundaries.
Second, if the available number of points is fixed for an object regardless of $n$, the density of $\pcd$ can change with different scales, \ie, splitting an object into $n^3$ voxels also splits input points into $1 / n^3$ per voxel.
Both can be easily mitigated by training with some iterations of $n\geq2$ and training coarser levels with a random number of input points.

\paragraph{Voxel subset training.}\noindent The encoder/decoder are applied independently to each subvoxel of the object. 
Thus, rather than training the entire object as would be required under standard token scaling, we can train just a subset of voxels, allowing for significantly reduced memory usage even when using many latent tokens. 
Note the forward pass still includes all active voxels (but without storing their activations), as we only have RGB renderings of full objects (not individual voxels). 

\paragraph{Implementation and ablation.}\noindent We build off LiTo to implement our encoder and decoder; see \Cref{sec:splitscaleimplementation} for details.
Following this above regimen, training comfortably fits within the memory limits of an 80GB GPU (\Cref{table:tokenizer_compute}) --- an important threshold given their relative availability (\eg, A100 and H100 GPUs) compared to GPUs with larger memory sizes. Simultaneously, compute time and FLOP usage remain comparable to a standard LiTo architecture. However, performance scales well even up to $131072$ total latent tokens ($n\approx8$), as visualized in \Cref{figure:tokenizer_curves}.

We ablate our design choices in downscaled experiments in \Cref{table:ablation_recon}, where we train tokenizers without various components of our final recipe for 44k iterations. Voxel normalization $\transform$ and training with both $n=1$ and $n=2$ can be seen to be crucial for performance, while training with $n=4$ and random point counts yield smaller improvements.

\begin{figure}[t]
    \centering
    \includegraphics[width=\linewidth]{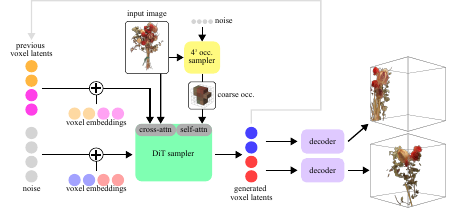} 
    \caption{\textbf{Visualization of our proposed generation procedure.} To reduce memory requirements, we propose autoregressive voxel-based generation with flow matching-based sampling and cross-attention conditioning. Given an input image, we start by first generating coarse $4^3$ occupancy. For the occupied voxels, we then autoregressively sample voxel latents conditioned on previously-generated latents, until all voxels are filled. Asset credit~\cite{dryflower}.
    }
    \label{figure:generator_method}
\end{figure}

\subsection{Autoregressive generation}

While split-and-scale provides a computationally-efficient avenue for training latent representations with a high number of latents, training a generative model for these larger latents is not straightforward. For instance, $n=4$ and $1024$ tokens per voxel translates to approximately 32768 total tokens on average --- in our experiments, naively scaling the LiTo diffusion transformer to generate 32768 tokens induces a peak training memory usage of nearly 160 GBs, far outpacing the capacity of most commonly available GPUs. 
To tackle this challenge, we propose an autoregressive voxel-based formulation with flow matching-based sampling~\cite{li2024autoregressive} and cross-attention conditioning, as visualized in \Cref{figure:generator_method}. 
In particular, we train a flow-matching model~\cite{lipman2022flow} based on a diffusion transformer~\cite{peebles2023scalable} that models the following distribution:
$
    P(\mathbf{v}_\text{target} \mid \mathbf{v}_\text{context}, O, I), 
$
where $\mathbf{v}_\text{target}$ denotes the latents of voxels to be generated, $\mathbf{v}_\text{context}$ the already-generated voxel latents, $O$ the $n^3$ coarse occupancy (as empty voxels are not reconstructed in split-and-scale), and $I$ the input image. 
We use self attention between noisy $\mathbf{v}_\text{target}$ tokens with cross attention to $\mathbf{v}_\text{context}$, mitigating the memory costs of larger latent generation. To apply other conditioning, we use cross attention with $I$ and self attention with the fewer token $O$.

\paragraph{Implementation and ablation.}\noindent We train this autoregressive model for split-and-scale latents from $n=4$ with $1024$ tokens per voxel, using randomly selected context and target voxels.
To minimize the overall number of autoregressive steps, we sample $8$ voxels at a time, resulting in $4$ flow-matching samplings per object on average. 
We train this model for $450$k iterations with a batch size of 256.
Note that training this model can again be performed on H100 80GB GPUs --- please see the appendix for more details on our network architecture, which mostly follows that of LiTo~\cite{lito}.

In \Cref{table:ablation_gen}, to evaluate our proposed autoregressive method, we also implement a naive approach that directly generates $n = 4$ split-and-scale latents with a DiT, but with reduced hidden dimension such that the model can be trained on an 80GB GPU. 
Our autoregressive approach can be seen to comfortably outperform this baseline when both models have been trained for 135k iterations.
For an equivalent amount of training time, our autoregressive model can also be seen to be competitive with a DiT trained to directly generate 32768 token LiTo latents, even though neither the representation nor the generative model fit within 80GB memory.

We also ablate potential inference voxel sampling orders in \Cref{table:ablation_gen}. 
Generating voxels in random orders versus in raster order away from the camera seems to behave comparably --- we use raster order for the results shown in this paper.
To leverage this model in practice, we also train a small flow-matching model that generates the $4\times4\times4$ occupancy of an object, conditioned on an input image. 
We ablate the impact of this additional generation step on overall performance in \Cref{table:ablation_gen}.

\begin{figure}[t]
    \centering
     \setlength{\tabcolsep}{0.5pt}
     \def\bgcolor{gray!80}
     \def\fgcolor{white}
     \begin{tabular}{ccccc|cc}
        \includegraphics[width=0.14\linewidth]{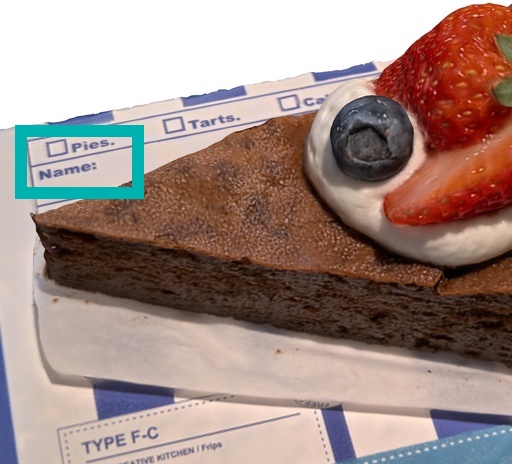} &  \includegraphics[width=0.14\linewidth]{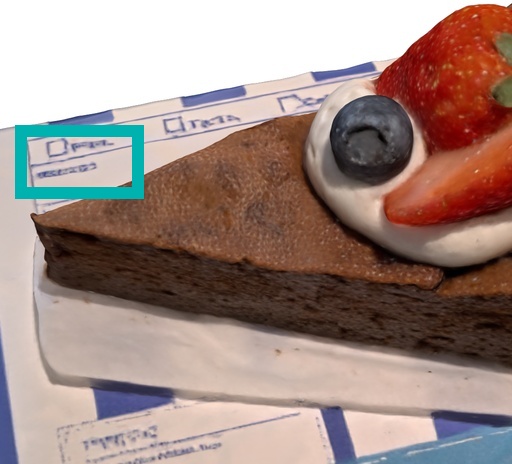} \put(-40, 45){\colorbox{\bgcolor}{\color{\fgcolor} \tiny LPIPS=0.24}} & 
        \includegraphics[width=0.14\linewidth]{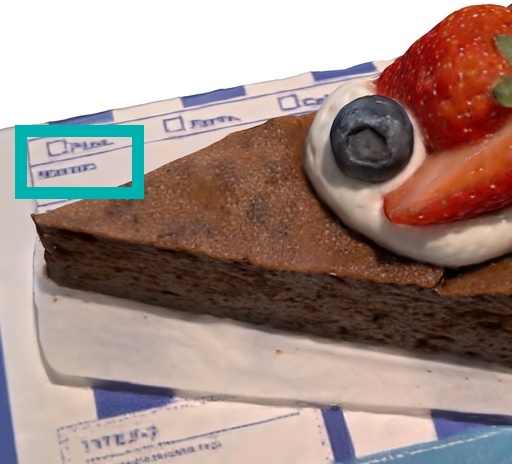} \put(-40, 45){\colorbox{\bgcolor}{\color{\fgcolor} \tiny LPIPS=0.22}} &  \includegraphics[width=0.14\linewidth]{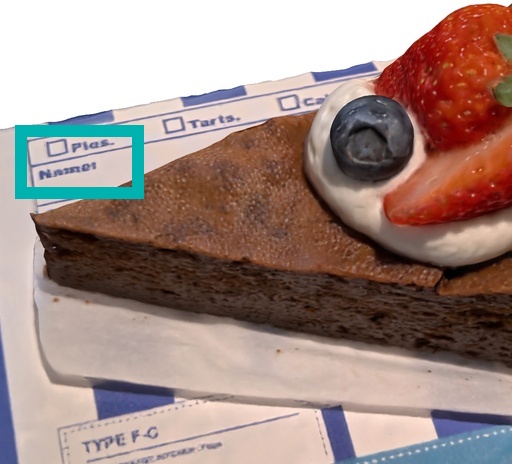} \put(-40, 45){\colorbox{\bgcolor}{\color{\fgcolor} \tiny LPIPS=0.16}}
        & \includegraphics[width=0.14\linewidth]{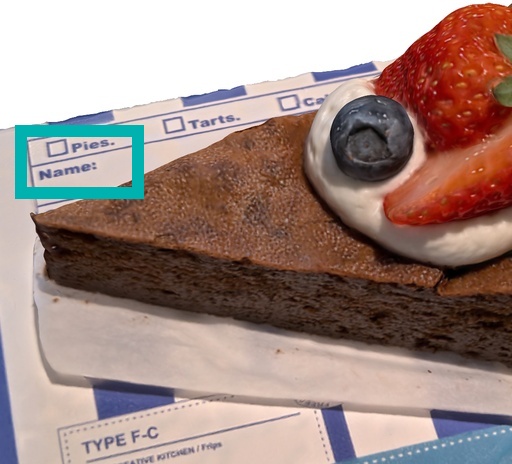} \put(-40, 45){\colorbox{\bgcolor}{\color{\fgcolor} \tiny LPIPS=0.08}} & \includegraphics[width=0.14\linewidth]{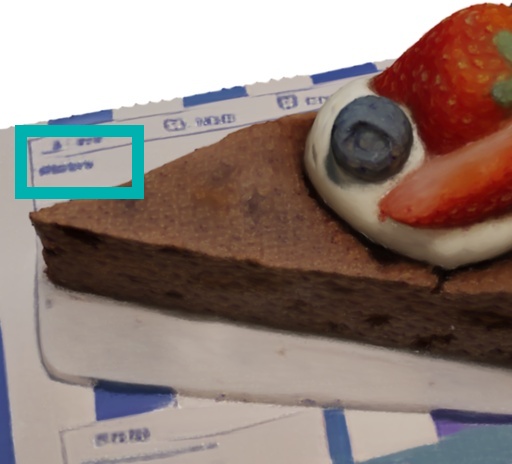} \put(-40, 45){\colorbox{\bgcolor}{\color{\fgcolor} \tiny LPIPS=0.36}}  & \includegraphics[width=0.14\linewidth]{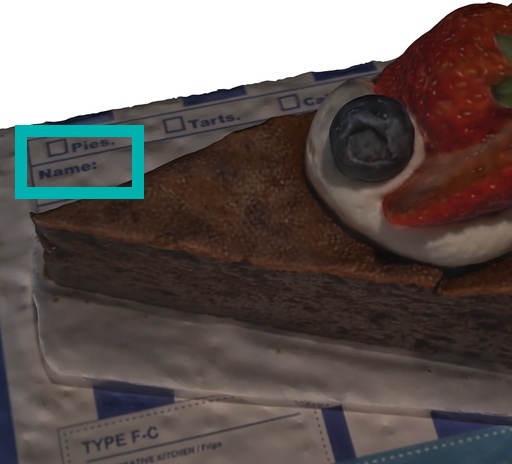} \put(-40, 45){\colorbox{\bgcolor}{\color{\fgcolor} \tiny LPIPS=0.28}}\\[-1ex]
        \includegraphics[width=0.14\linewidth]{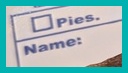} & \includegraphics[width=0.14\linewidth]{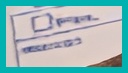} & \includegraphics[width=0.14\linewidth]{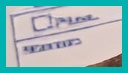} &  \includegraphics[width=0.14\linewidth]{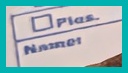}  
        & \includegraphics[width=0.14\linewidth]{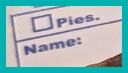}  & \includegraphics[width=0.14\linewidth]{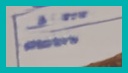}  & \includegraphics[width=0.14\linewidth]{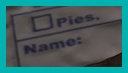}\\
        \includegraphics[width=0.14\linewidth]{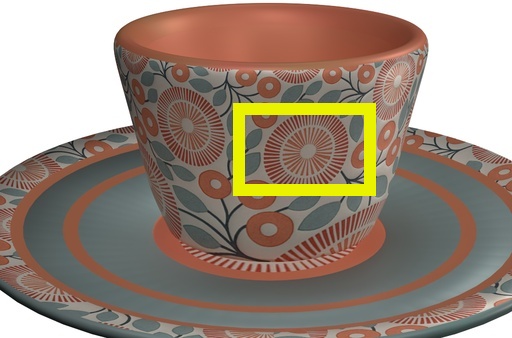} & \includegraphics[width=0.14\linewidth]{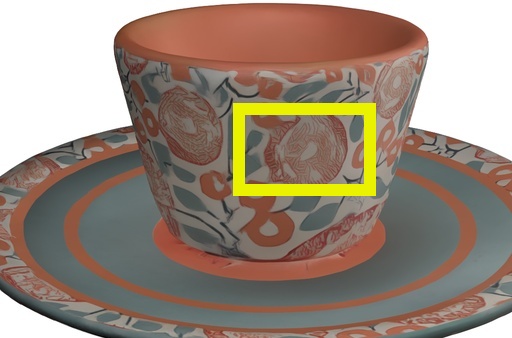} \put(-40, 40){\colorbox{\bgcolor}{\color{\fgcolor} \tiny LPIPS=0.15}} & \includegraphics[width=0.14\linewidth]{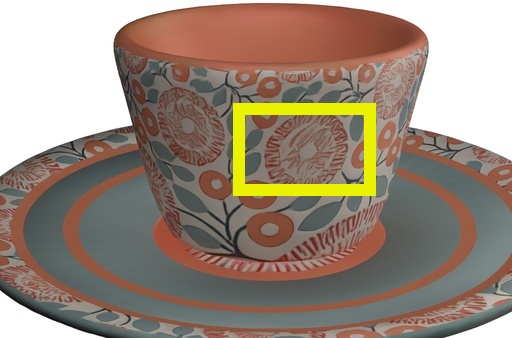} \put(-40, 40){\colorbox{\bgcolor}{\color{\fgcolor} \tiny LPIPS=0.10}} & \includegraphics[width=0.14\linewidth]{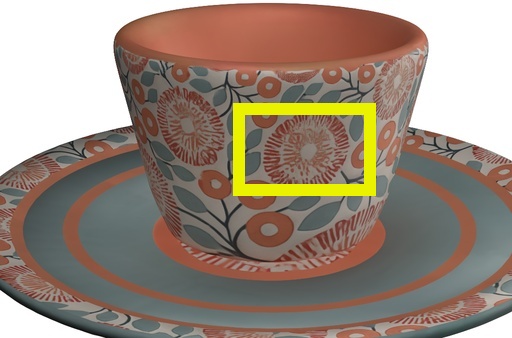} \put(-40, 40){\colorbox{\bgcolor}{\color{\fgcolor} \tiny LPIPS=0.08}}&
        \includegraphics[width=0.14\linewidth]{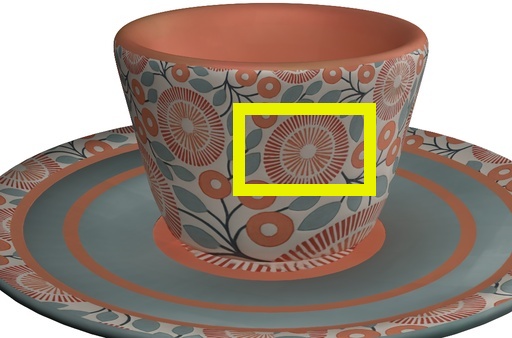} \put(-40, 40){\colorbox{\bgcolor}{\color{\fgcolor} \tiny LPIPS=0.04}} &\includegraphics[width=0.14\linewidth]{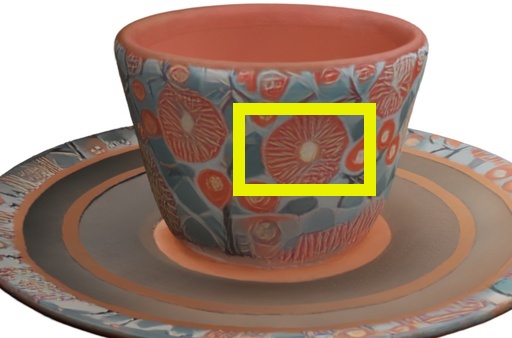} \put(-40, 40){\colorbox{\bgcolor}{\color{\fgcolor} \tiny LPIPS=0.27}} & \includegraphics[width=0.14\linewidth]{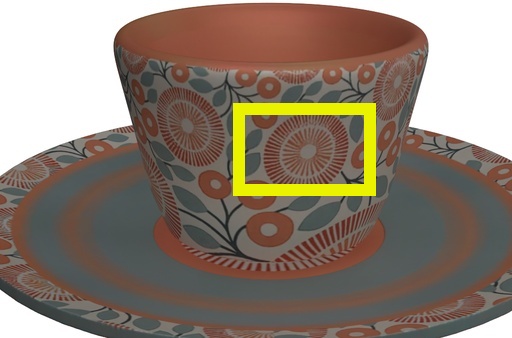} \put(-40, 40){\colorbox{\bgcolor}{\color{\fgcolor} \tiny LPIPS=0.14}} \\[-1ex]
        \includegraphics[width=0.14\linewidth]{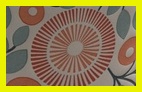} & \includegraphics[width=0.14\linewidth]{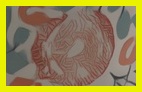} & \includegraphics[width=0.14\linewidth]{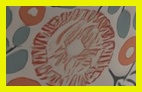} & \includegraphics[width=0.14\linewidth]{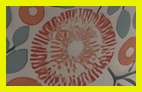}
        & \includegraphics[width=0.14\linewidth]{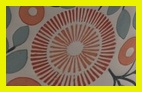}&\includegraphics[width=0.14\linewidth]{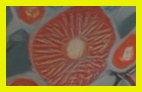} & \includegraphics[width=0.14\linewidth]{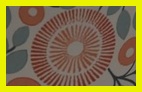} \\
        \includegraphics[width=0.14\linewidth]{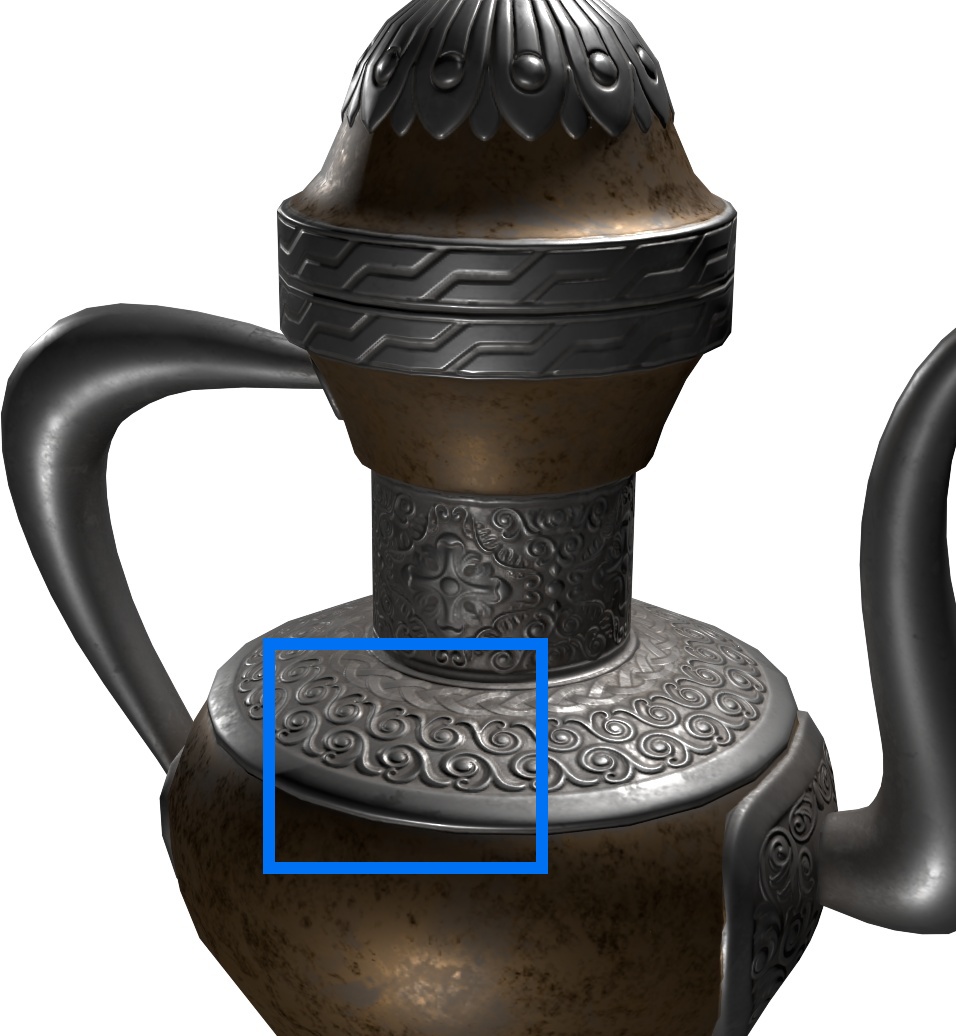} & \includegraphics[width=0.14\linewidth]{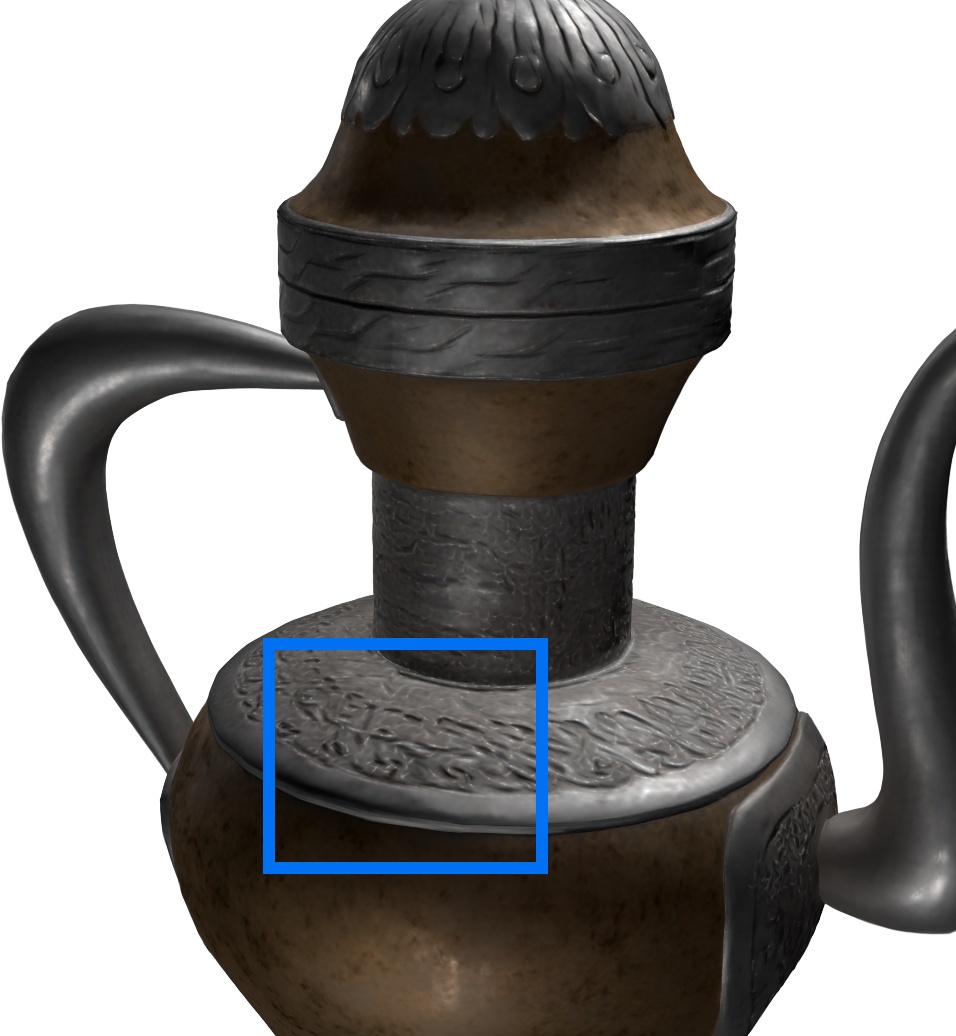} \put(-40, 64){\colorbox{\bgcolor}{\color{\fgcolor} \tiny LPIPS=0.17}} & \includegraphics[width=0.14\linewidth]{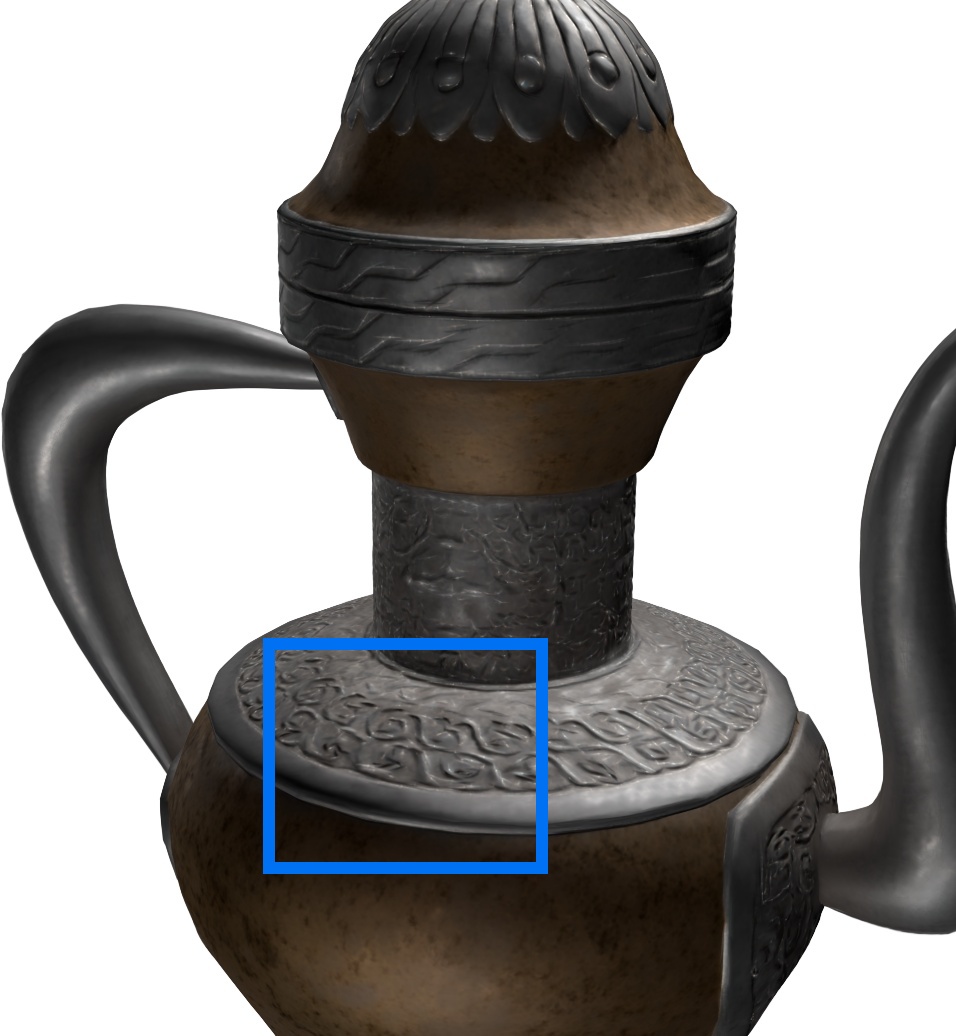} \put(-40, 64){\colorbox{\bgcolor}{\color{\fgcolor} \tiny LPIPS=0.15}} & \includegraphics[width=0.14\linewidth]{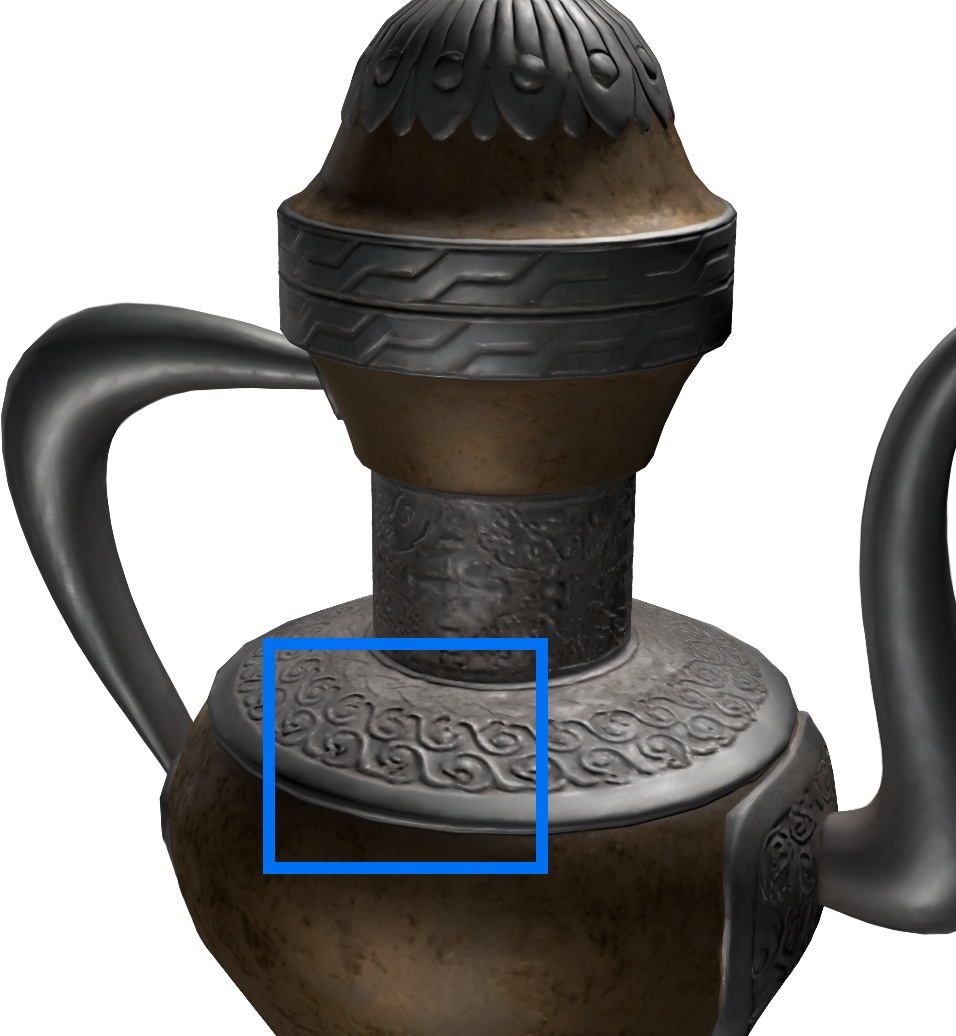} \put(-40, 64){\colorbox{\bgcolor}{\color{\fgcolor} \tiny LPIPS=0.13}} &

        \includegraphics[width=0.14\linewidth]{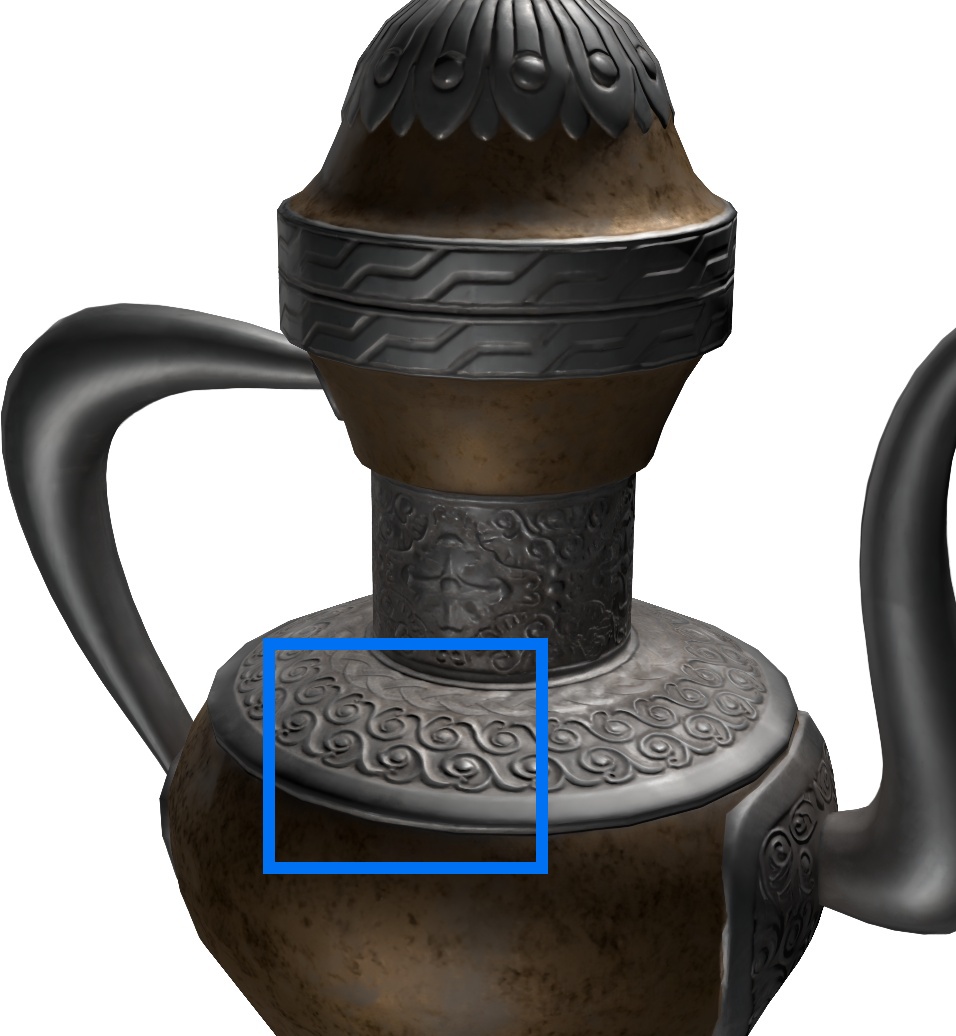} \put(-40, 64){\colorbox{\bgcolor}{\color{\fgcolor} \tiny LPIPS=0.09}} & \includegraphics[width=0.14\linewidth]{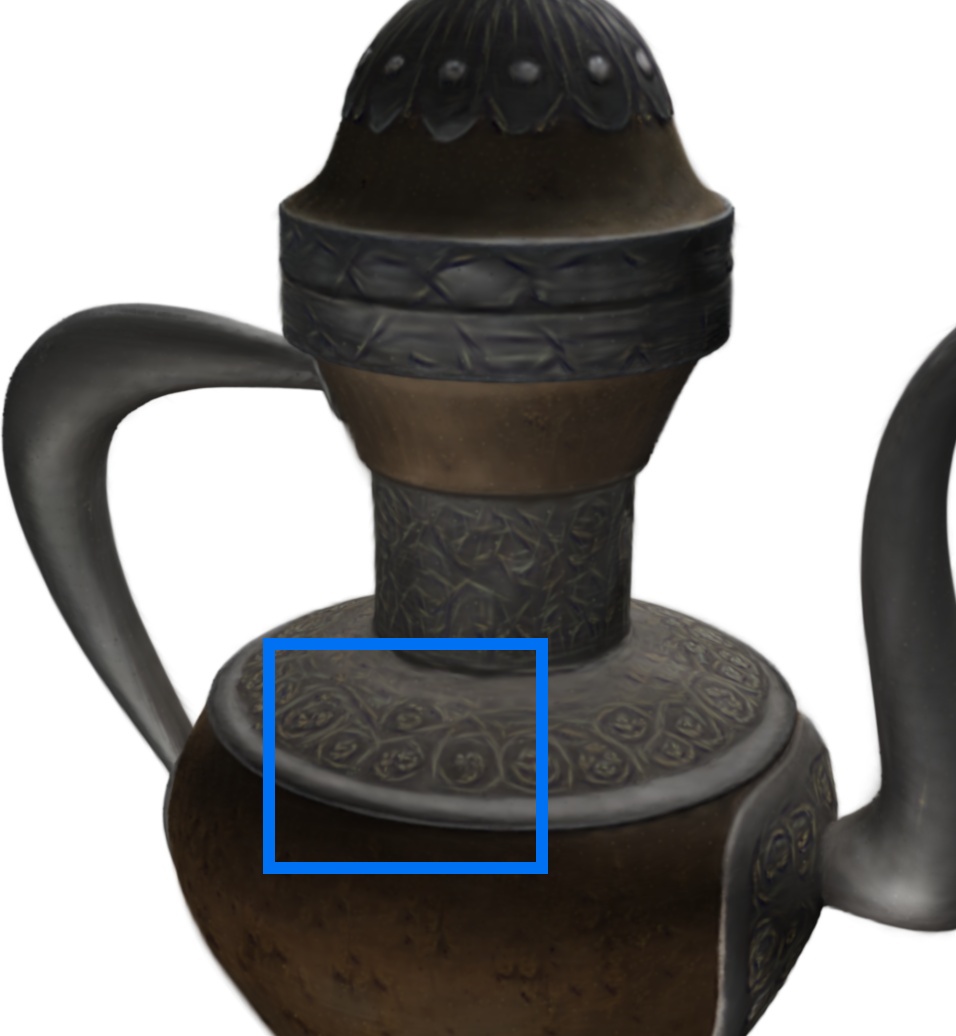} \put(-40, 64){\colorbox{\bgcolor}{\color{\fgcolor} \tiny LPIPS=0.25}} & \includegraphics[width=0.14\linewidth]{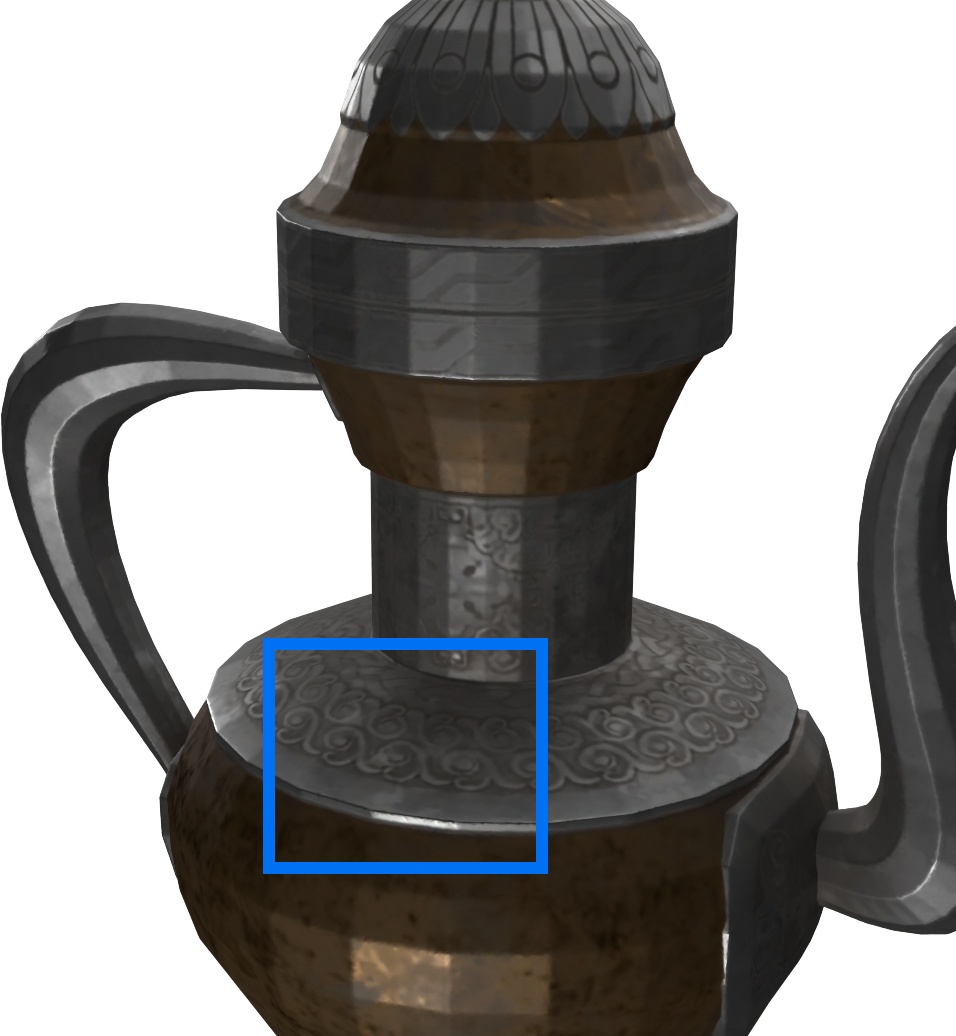} \put(-40, 64){\colorbox{\bgcolor}{\color{\fgcolor} \tiny LPIPS=0.21}} \\[-1ex]
        \includegraphics[width=0.14\linewidth]{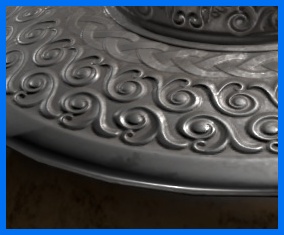} & \includegraphics[width=0.14\linewidth]{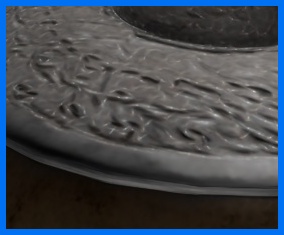} & \includegraphics[width=0.14\linewidth]{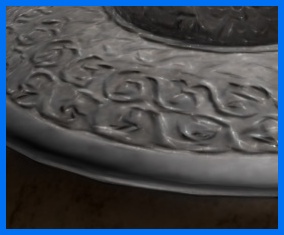} & \includegraphics[width=0.14\linewidth]{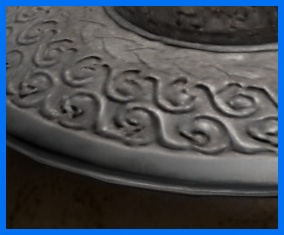} &
        \includegraphics[width=0.14\linewidth]{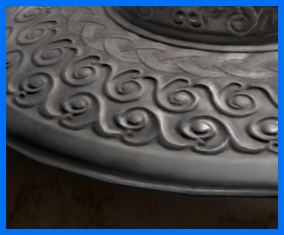} & \includegraphics[width=0.14\linewidth]{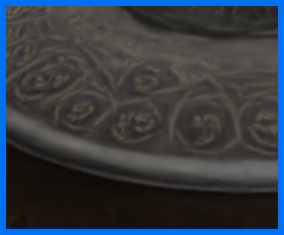} & \includegraphics[width=0.14\linewidth]{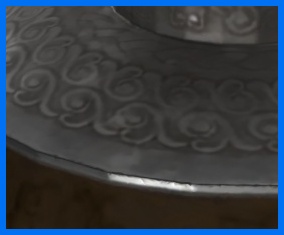}\\
        \footnotesize GT & \begin{tabular}{c}\footnotesize LiTo-8192\\\footnotesize (8192 inf.)\end{tabular} & \begin{tabular}{c}\footnotesize LiTo-8192 \\\footnotesize (32768 inf.)\end{tabular} &
        \begin{tabular}{c}\textit{\footnotesize Split-and-scale} \\\footnotesize (32768 inf.)\end{tabular} 
        & \begin{tabular}{c}\footnotesize \textit{Split-and-scale} \\\footnotesize (131072 inf.)\end{tabular} & \footnotesize TRELLIS & \footnotesize TRELLIS-2
    \end{tabular}
    \caption{\textbf{Qualitative reconstruction comparison.} For comparison, we visualize a LiTo model trained with 8192 tokens (``LiTo-8192''), that can be trained under similar compute limits as our split-and-scale model. 
    Note that the metrics for TRELLIS and TRELLIS-2 are quite poor due to color shift. Asset credits~\cite{cake,teacup,jug}.}
    \label{figure:tokenizer_qual}
\end{figure}

\section{Results}

\subsection{Reconstruction}
\label{sec:recon}
\paragraph{Dataset.}\noindent We build our split-and-scale representation off TRELLIS's~\cite{xiang2025trellis} training subset of Objaverse-XL~\cite{deitke2023objaversexl}, which we split into 10\% validation, 10\% test and 80\% training sets following LiTo~\cite{lito}. 
For quantitative evaluation, we randomly select 200 objects from the 1000 test set objects with highest aesthetic score as computed by TRELLIS.\footnote{This means our test set is TRELLIS's and TRELLIS-2's training set.} 
We follow a simplified LiTo pipeline and render 1036$\times$1036 resolution images under random lighting. 
To get the input point cloud, we backproject 150 views rendered on a sphere and randomly select 16 million points. 
We supervise the Gaussian decoder with 100 images rendered from random views.

\paragraph{Results.}\noindent 
For quantitative evaluation, we first compare our split-and-scale model with baseline LiTo models~\cite{lito} trained for a variety of different latent counts, under both equal iteration~(\Cref{figure:tokenizer_curves}) and equal compute time~(\Cref{figure:tokenizer_curves_full}(a)) settings. 
In both cases, it can be seen that for an equivalent total latent count, split-and-scale is competitive with a LiTo model trained with that total latent count. 
Additionally, split-and-scale outperforms the comparable LiTo model when scaled to use more tokens, \eg, split-and-scale with $4\times4\times4$ voxels ($\approx32768$ tokens) performs better than a LiTo model trained to use 8192 tokens, even when that LiTo model is applied with more latents at inference. 
For completeness, we also include a 32768 token variant of LiTo (LiTo-32768), that notably cannot be trained on an H100 unlike the other methods shown. 
Nevertheless, split-and-scale still performs comparably. 
We also evaluate the geometry quality via Chamfer distance between ground truth point clouds and point clouds sampled from the flow-matching decoder of LiTo and split-and-scale. 
As shown in \Cref{figure:geometry_curves}, point clouds sampled from split-and-scale can be seen to significantly outperform those sampled from comparable LiTo models.

In \Cref{table:baseline_table}, we compare with other baselines more broadly (we include our LiTo implementation trained with 8192 tokens, \ie, ``LiTo-8192'', for completeness). 
To compare with the geometry of mesh-based approaches, \eg TRELLIS~\cite{xiang2025trellis} or TRELLIS-2~\cite{xiang2025trellis2}, we also train a mesh decoder from our split-and-scale representation (network architecture discussed in appendix). 
Our approach can be seen to be broadly competitive across both appearance and geometry.

For qualitative evaluation, in \Cref{figure:tokenizer_qual} and the supplemental website, we visualize the results of split-and-scale relative to LiTo-8192 as well as recent baselines~\cite{xiang2025trellis,xiang2025trellis2}. 
As can be seen, split-and-scale performs well in regions of high detail compared to a computationally-similar LiTo model trained with 8192 tokens.
Simultaneously, it lacks the color drift issues of TRELLIS~\cite{xiang2025trellis} and TRELLIS-2~\cite{xiang2025trellis2}\footnote{This color shift occurs despite rendering TRELLIS-2's textured mesh outputs with ground truth lighting (\Cref{sec:implementation}).}. We also show a comparison with the upper-bound LiTo-32768 in \Cref{figure:upper_tokenizer_qual}.

\begin{table}[h]
    \centering
     \caption{\textbf{Quantitative reconstruction comparison with other baselines.}}
    \label{table:baseline_table}
    \adjustbox{max width=0.9\columnwidth}{%
    \begin{tabular}{lccccc}
        \toprule
        \textbf{Method} & \textbf{Infer. latent size} & \textbf{LPIPS} $\downarrow$ & \textbf{Chamfer (mesh) $\downarrow$} & \textbf{Chamfer (flow)} $\downarrow$\\
        \midrule
        LiTo-8192 & $8192\times32$ & 0.124 & - & $2.87\times10^{-3}$\\
        LiTo-8192 & $32768\times32$ & \second 0.106 & - & $2.74\times10^{-3}$ \\
        TRELLIS & $\approx20\text{k}\times11$ & 0.230 & $1.28\times10^{-2}$ & -\\
        TRELLIS-2 & $\approx10\text{k}\times32$ & 0.184 & $1.53\times10^{-2}$ & -\\
        \midrule
        \textit{Split-and-scale} & $8192\times32$ & 0.124 & \second $1.70\times10^{-3}$ & \second $2.46\times10^{-3}$ \\
        \textit{Split-and-scale} & $32768\times32$ & \first 0.091 & \first $1.57\times10^{-3}$ & \first $2.09\times10^{-3}$\\
        \bottomrule
    \end{tabular}%
    }
\end{table}

\begin{figure}[ht!]
    \centering
     \setlength{\tabcolsep}{0.5pt}
    \begin{tabular}{c}
        \begin{tabular}{ccc}
            \fbox{\includegraphics[height=0.15\linewidth]{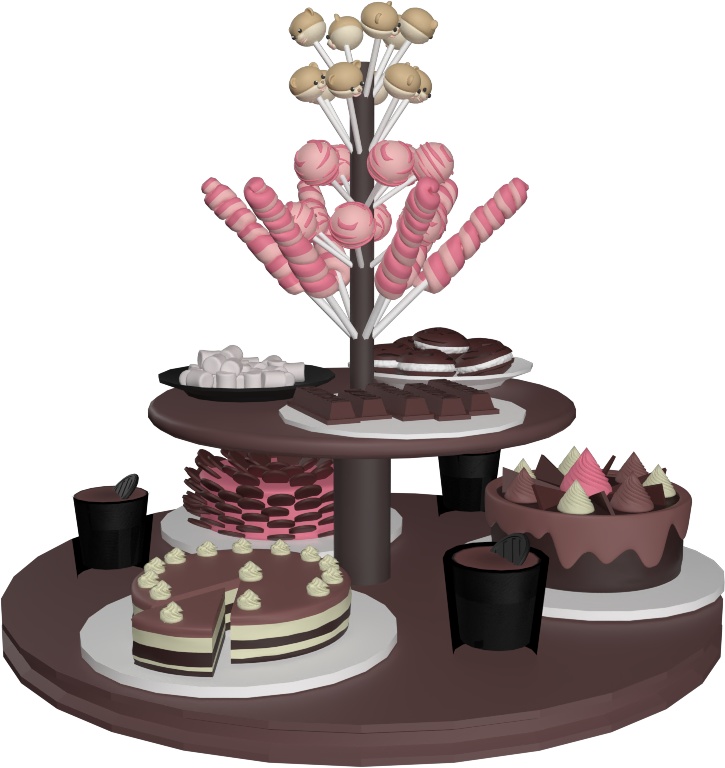}} & \fbox{\includegraphics[height=0.15\linewidth]{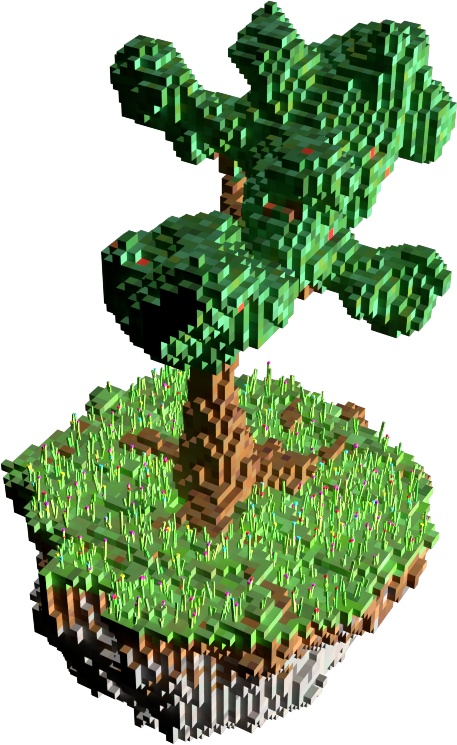}} & 
            \fbox{\includegraphics[height=0.15\linewidth]{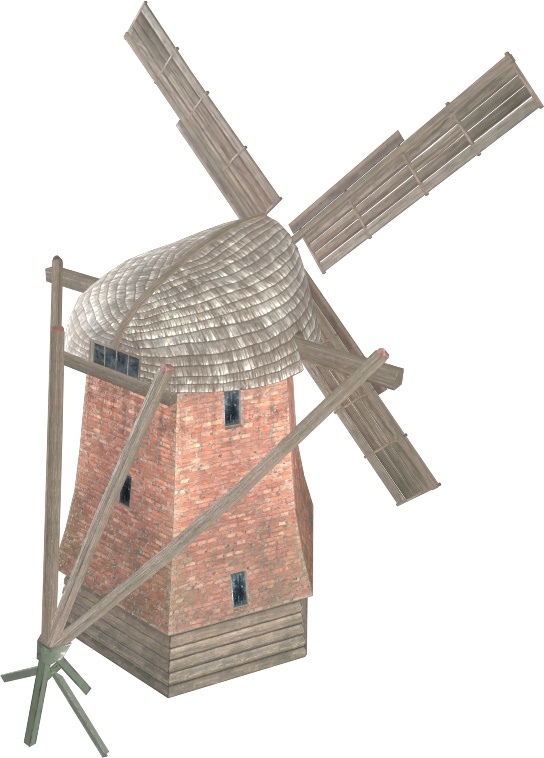}}\\
            \multicolumn{3}{c}{Input images}
        \end{tabular} \\
        \begin{tabular}{cccccccc}
            \includegraphics[width=0.123\linewidth]{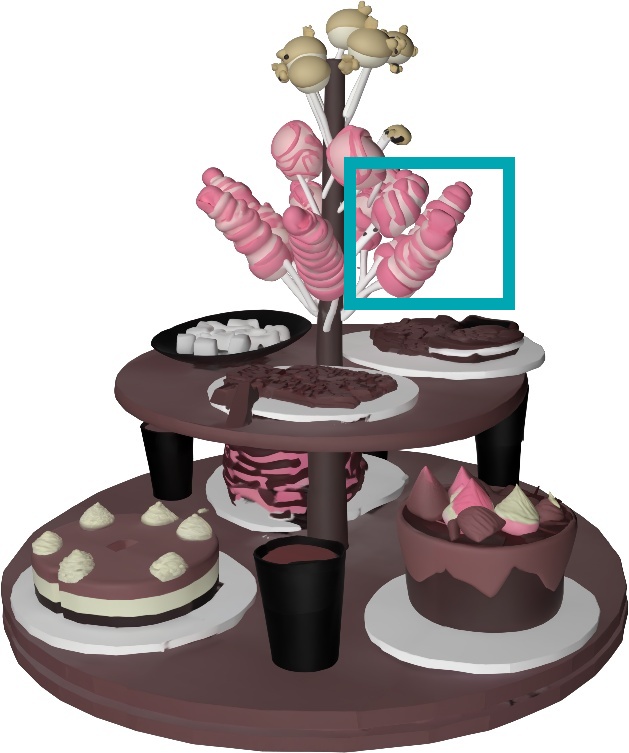} & \includegraphics[width=0.123\linewidth]{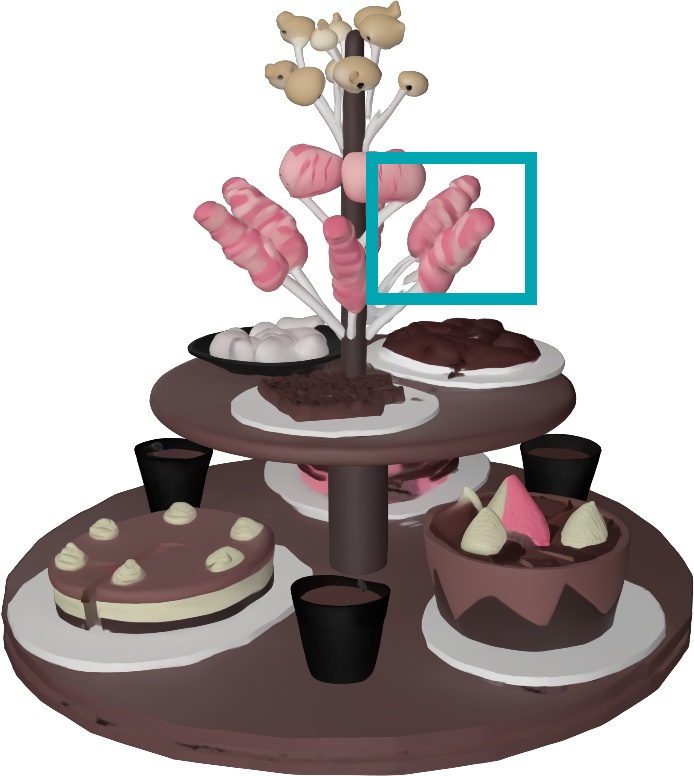} & \includegraphics[width=0.123\linewidth]{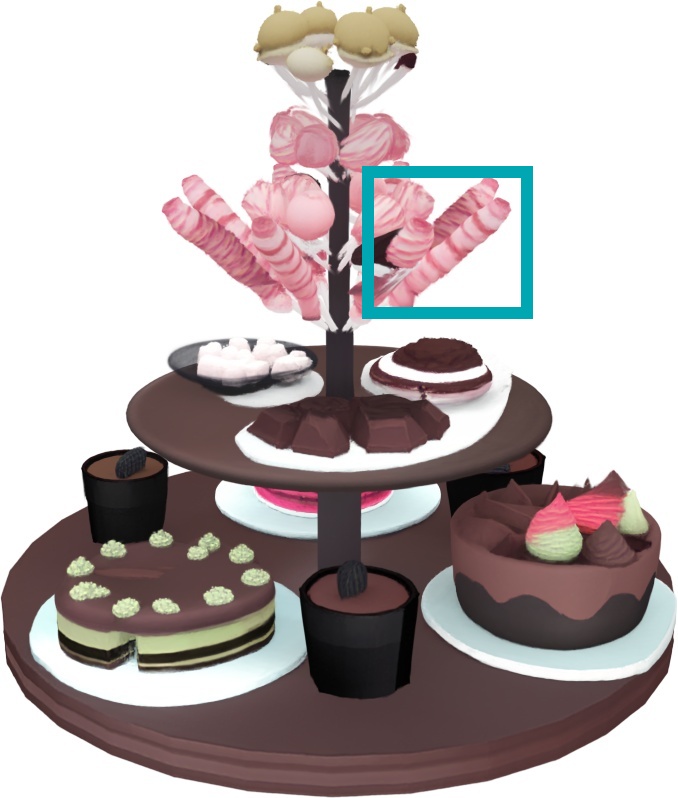} & \includegraphics[width=0.123\linewidth]{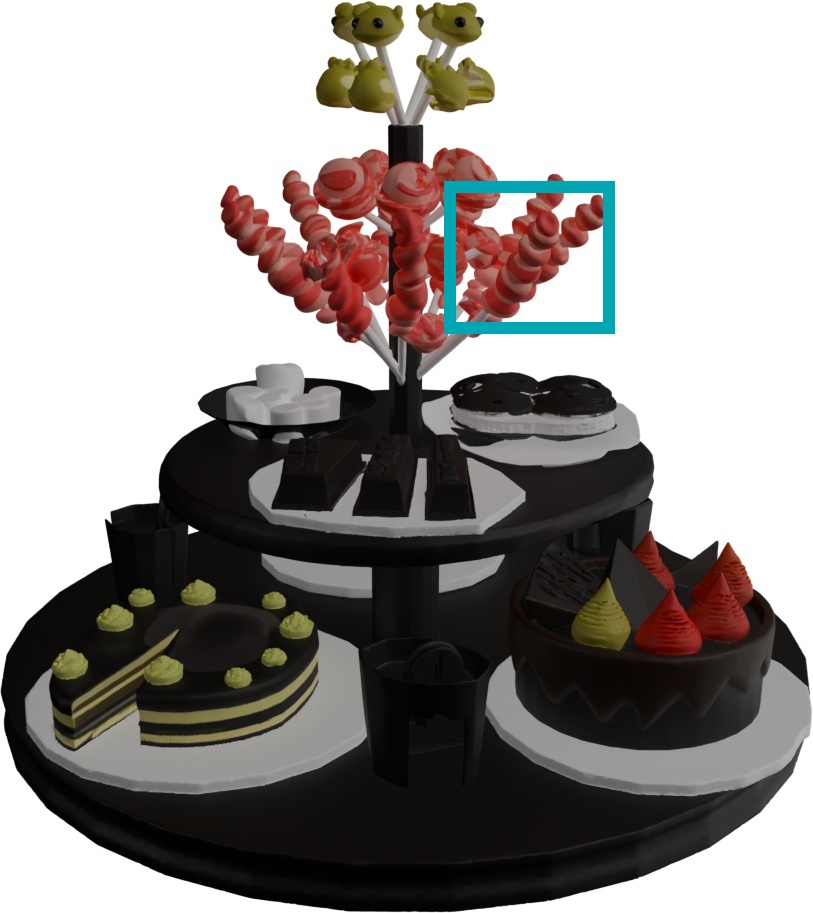} &\includegraphics[width=0.123\linewidth]{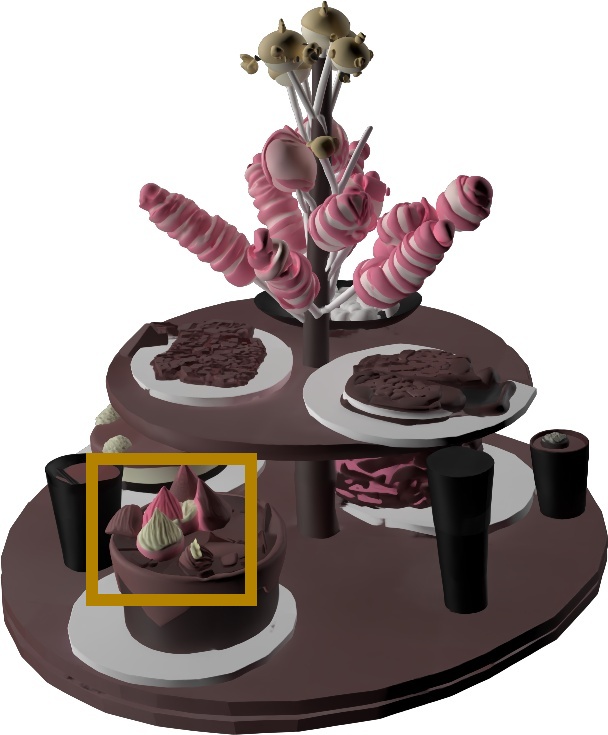} & \includegraphics[width=0.123\linewidth]{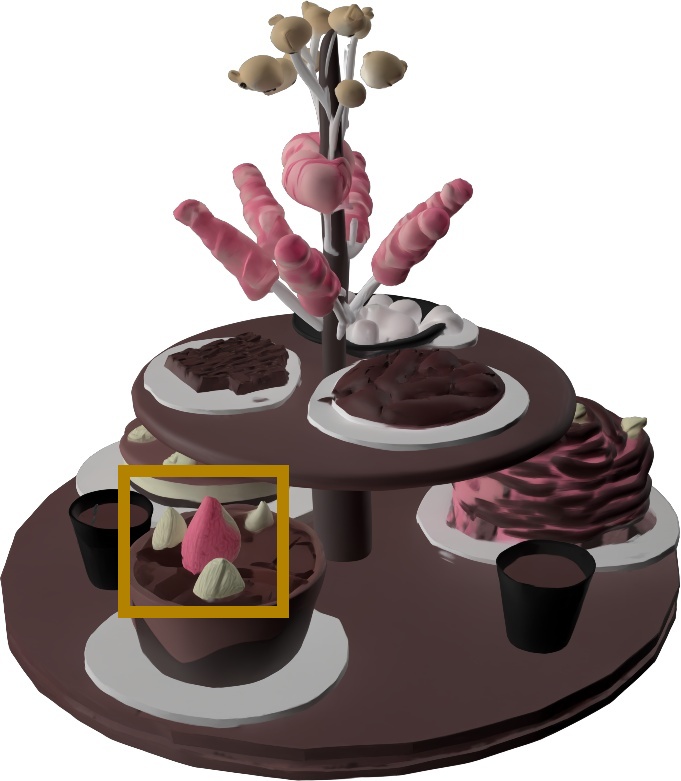}  & \includegraphics[width=0.123\linewidth]{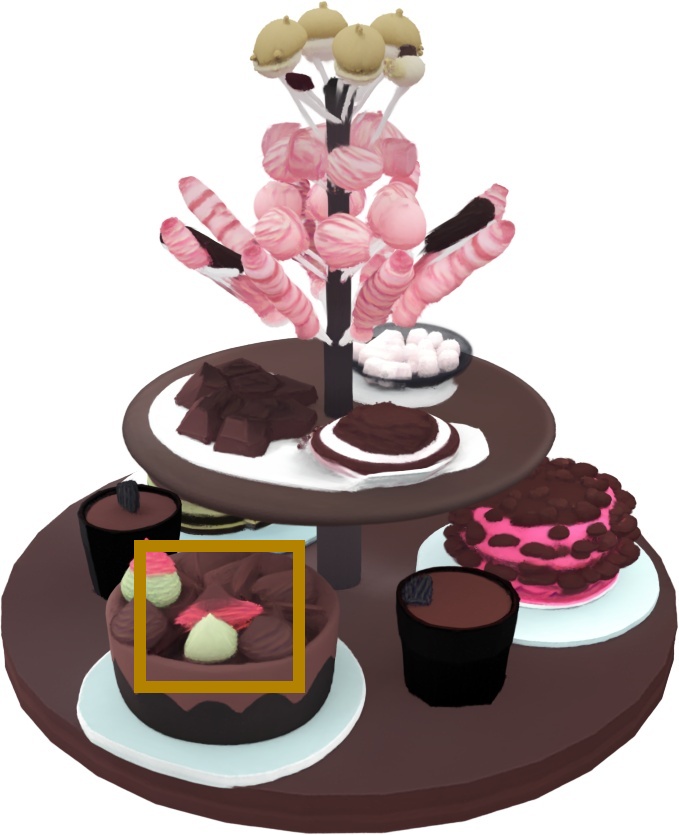} & \includegraphics[width=0.123\linewidth]{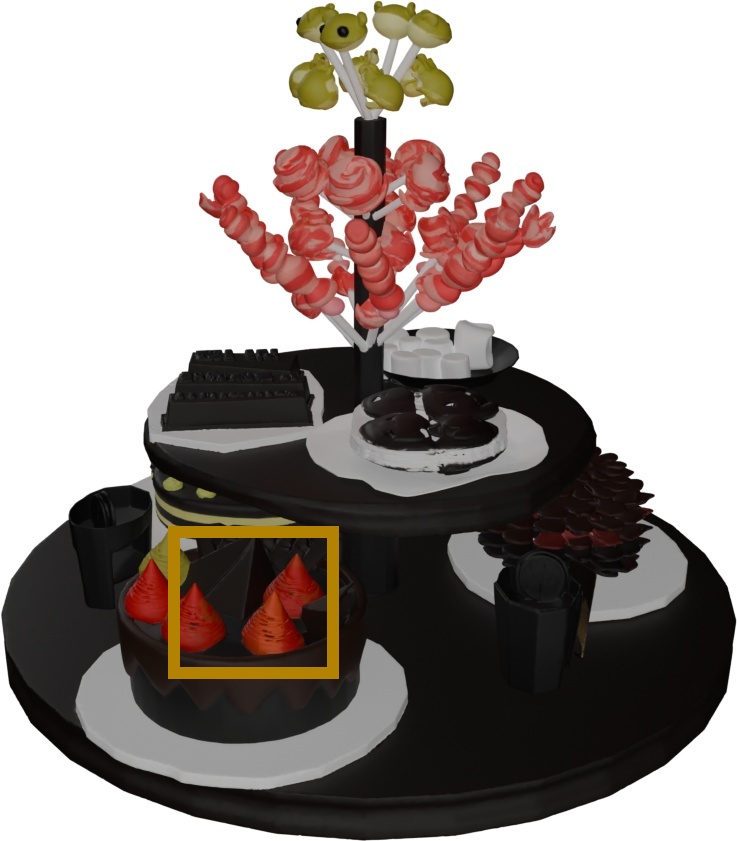}
            
            \\
            \includegraphics[width=0.123\linewidth]{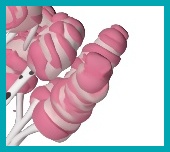} & \includegraphics[width=0.123\linewidth]{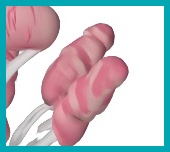} & \includegraphics[width=0.123\linewidth]{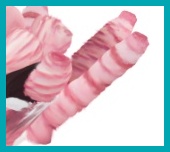}
            & \includegraphics[width=0.123\linewidth]{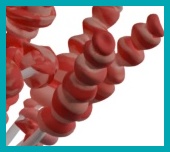}
            & \includegraphics[width=0.123\linewidth]{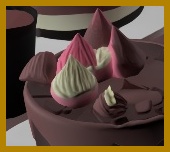} & \includegraphics[width=0.123\linewidth]{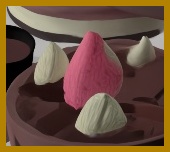} & \includegraphics[width=0.123\linewidth]{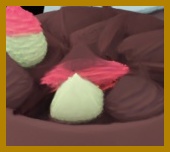}
            & \includegraphics[width=0.123\linewidth]{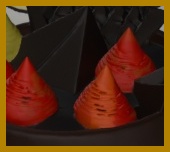}
            \\
            \includegraphics[width=0.123\linewidth]{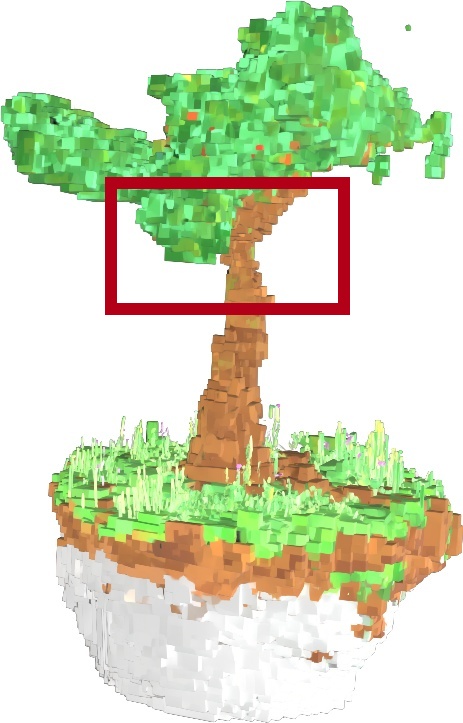} & \includegraphics[width=0.123\linewidth]{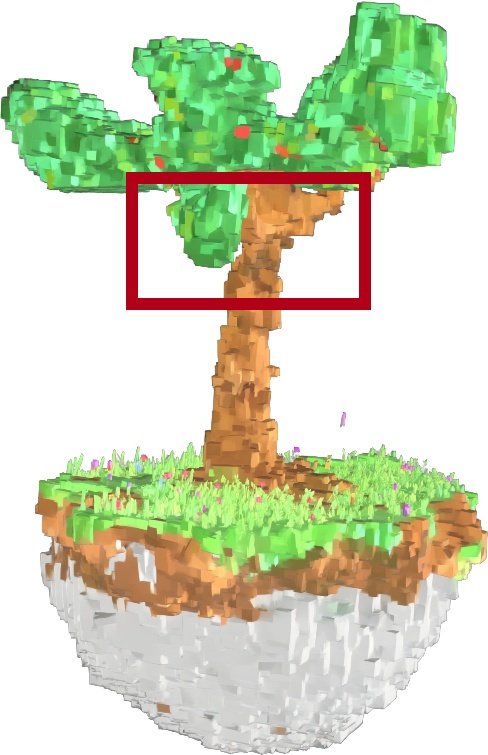} & \includegraphics[width=0.123\linewidth]{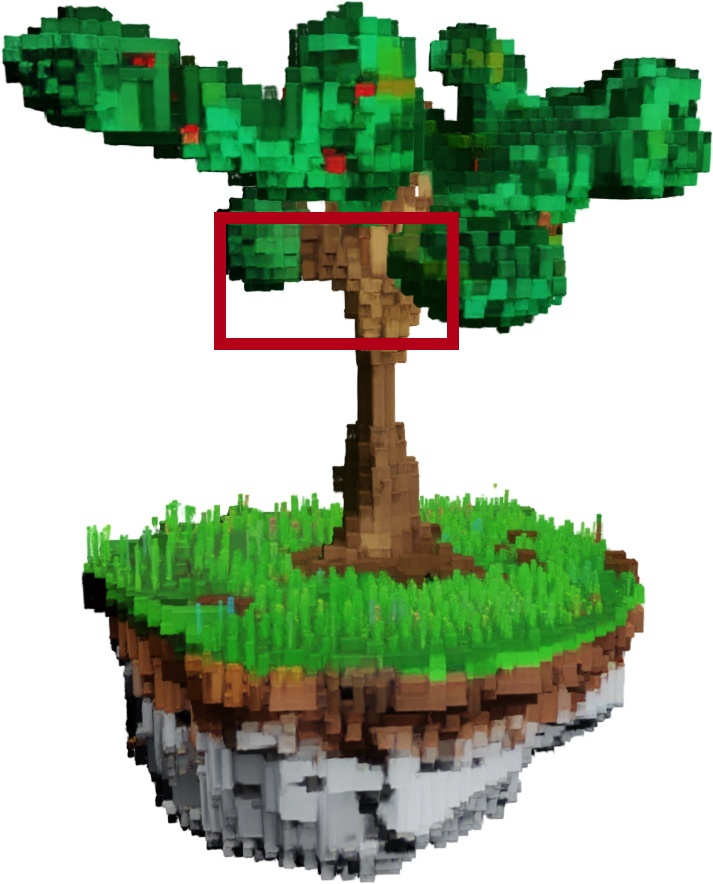} & \includegraphics[width=0.123\linewidth]{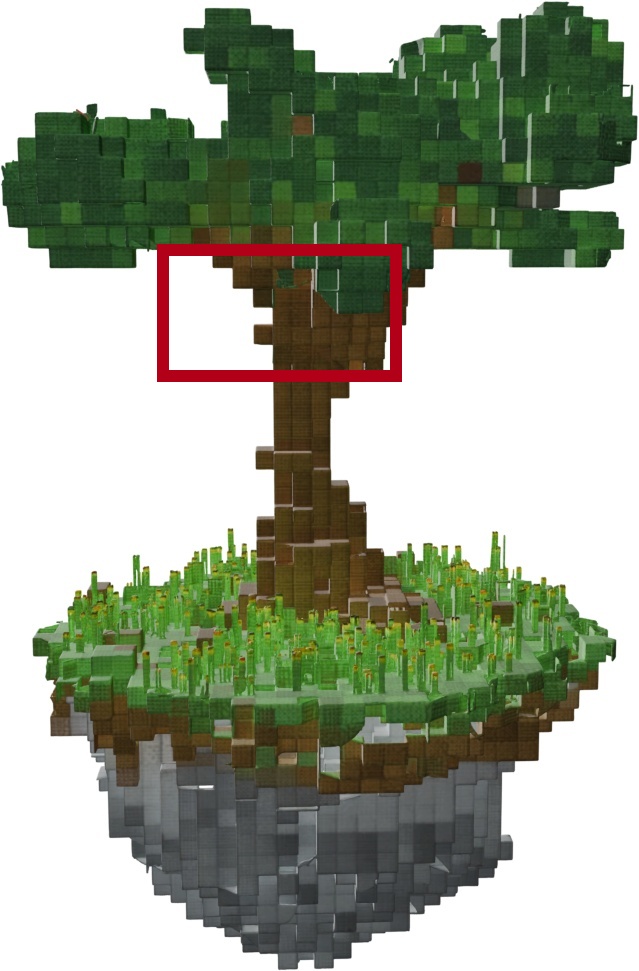} & \includegraphics[width=0.123\linewidth]{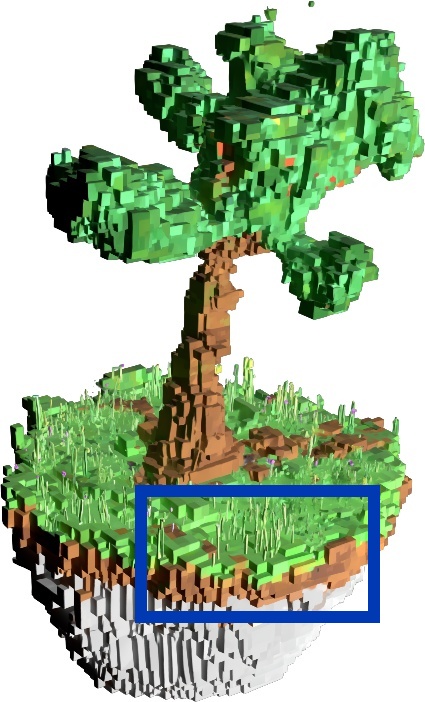} & \includegraphics[width=0.123\linewidth]{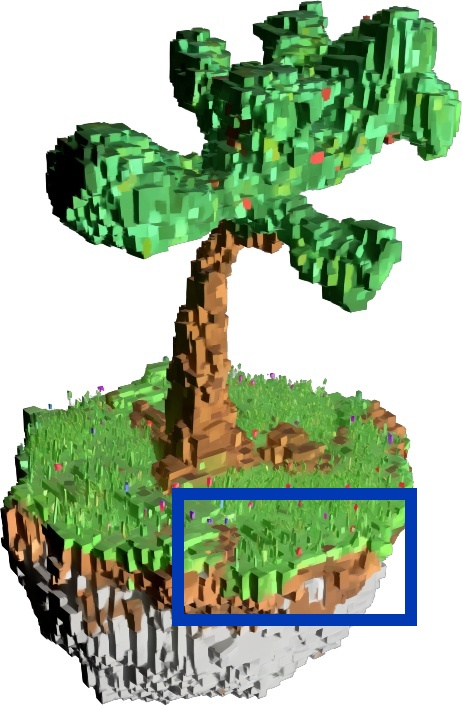}& \includegraphics[width=0.123\linewidth]{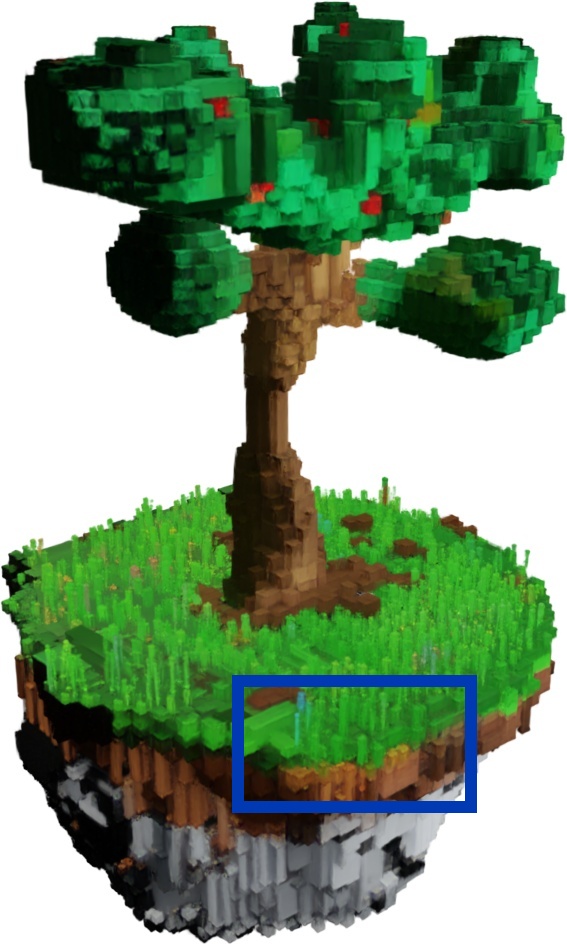} & \includegraphics[width=0.123\linewidth]{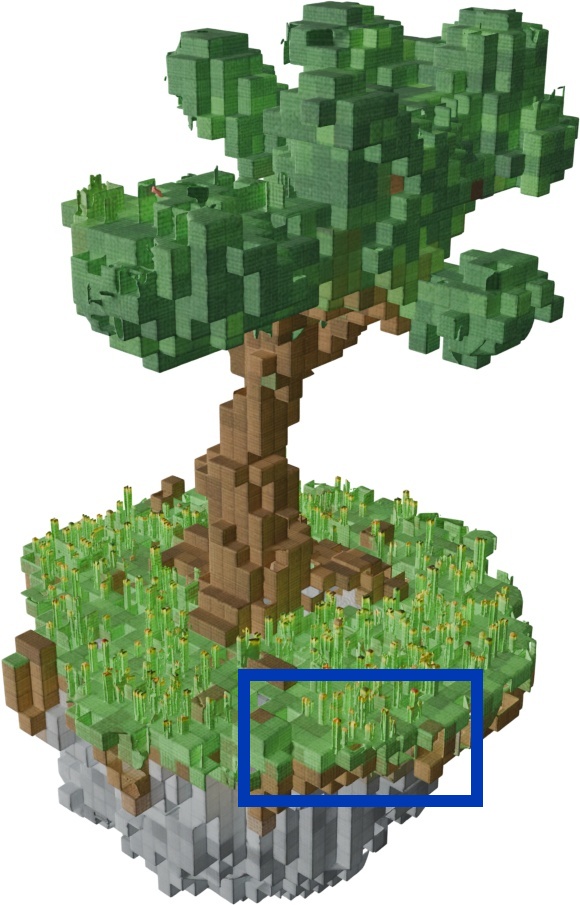}\\
            \includegraphics[width=0.123\linewidth]{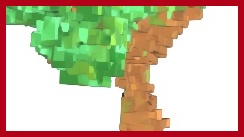} & \includegraphics[width=0.123\linewidth]{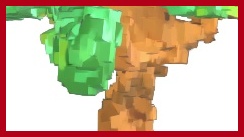} & \includegraphics[width=0.123\linewidth]{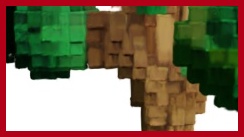} & \includegraphics[width=0.123\linewidth]{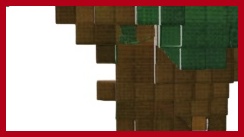} & \includegraphics[width=0.123\linewidth]{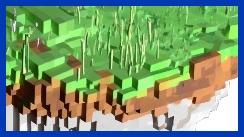} & \includegraphics[width=0.123\linewidth]{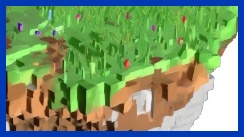} & \includegraphics[width=0.123\linewidth]{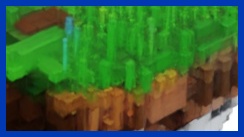} & \includegraphics[width=0.123\linewidth]{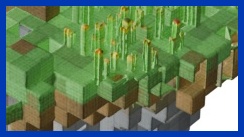}\\

            \includegraphics[width=0.123\linewidth]{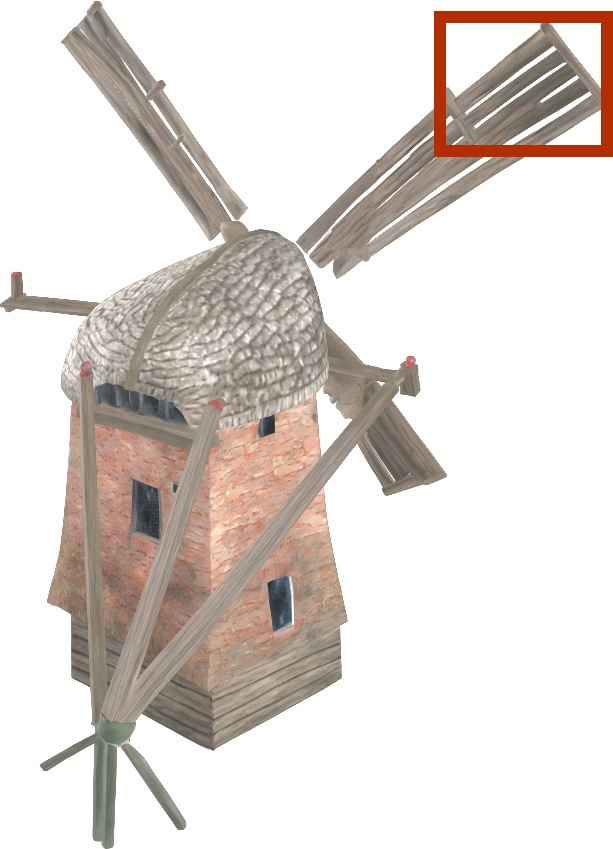} & \includegraphics[width=0.123\linewidth]{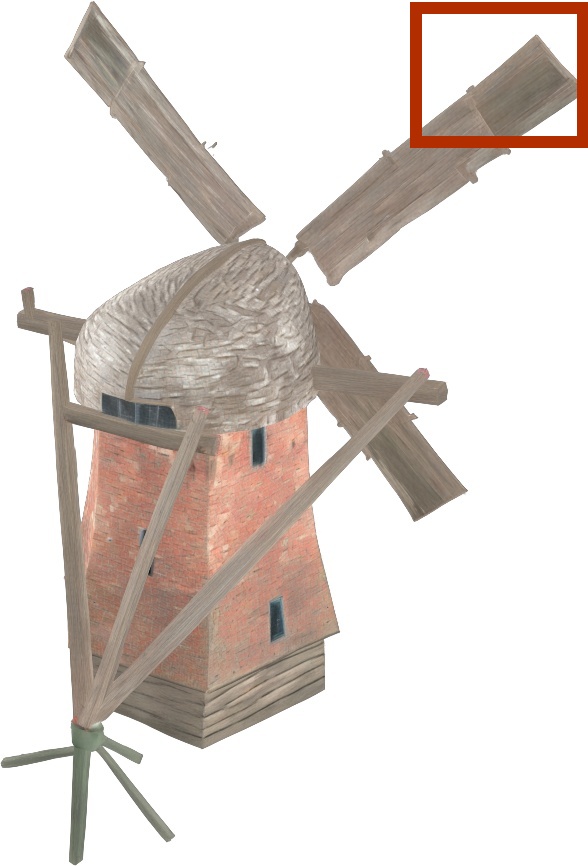} & \includegraphics[width=0.123\linewidth]{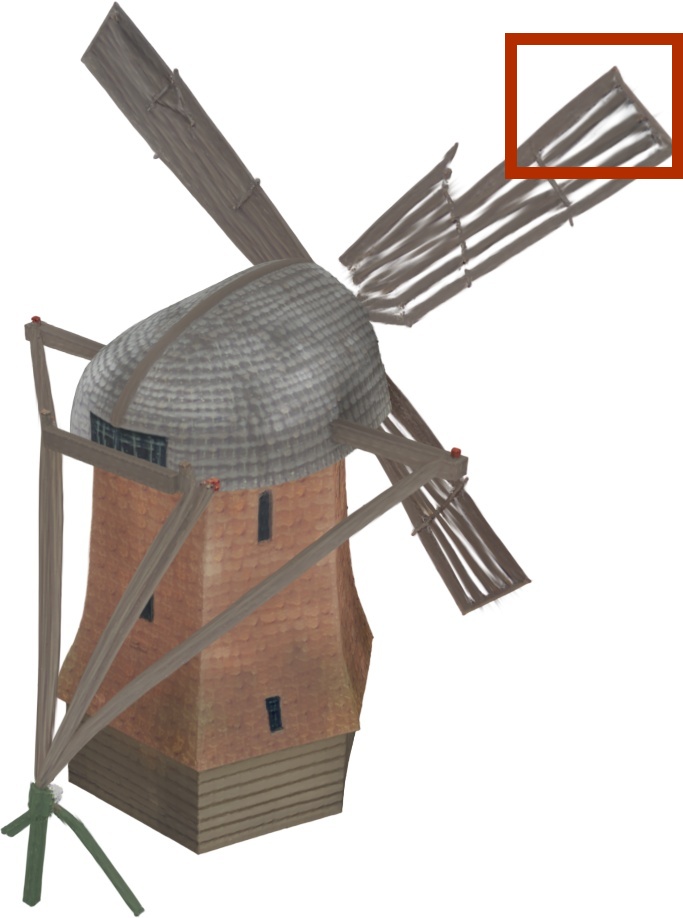} & \includegraphics[width=0.123\linewidth]{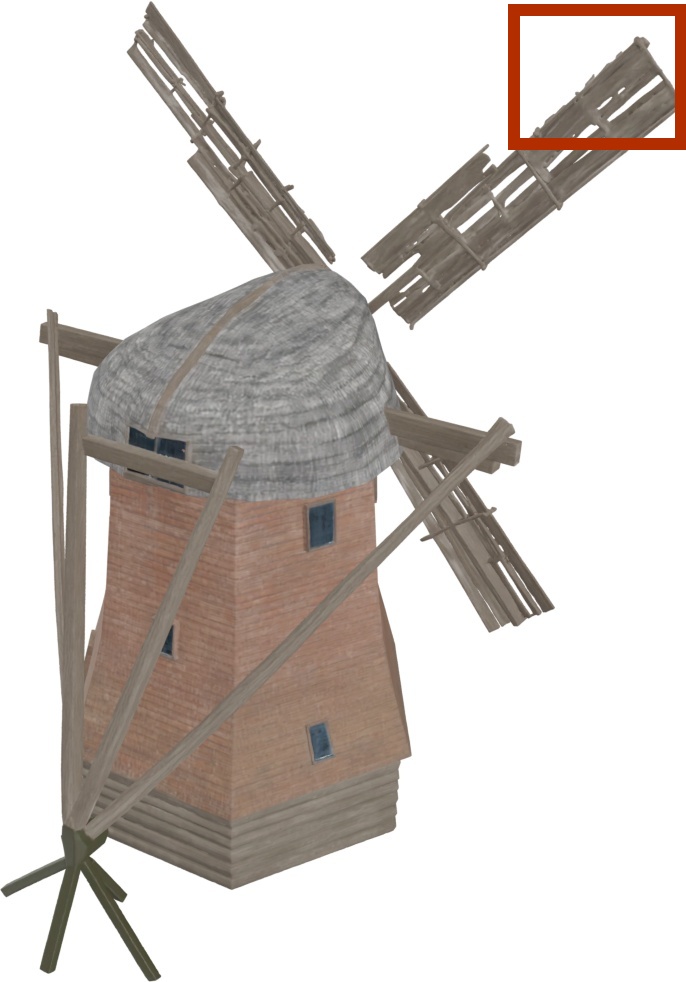} &  \includegraphics[width=0.123\linewidth]{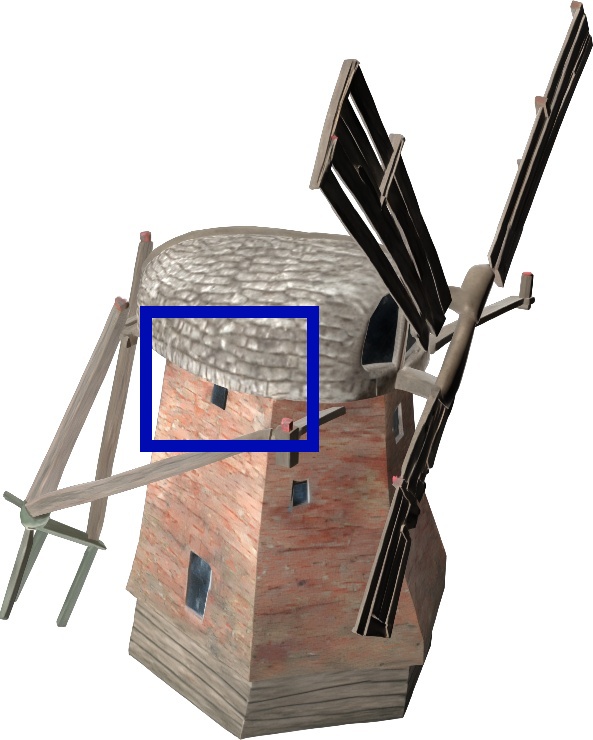} & \includegraphics[width=0.123\linewidth]{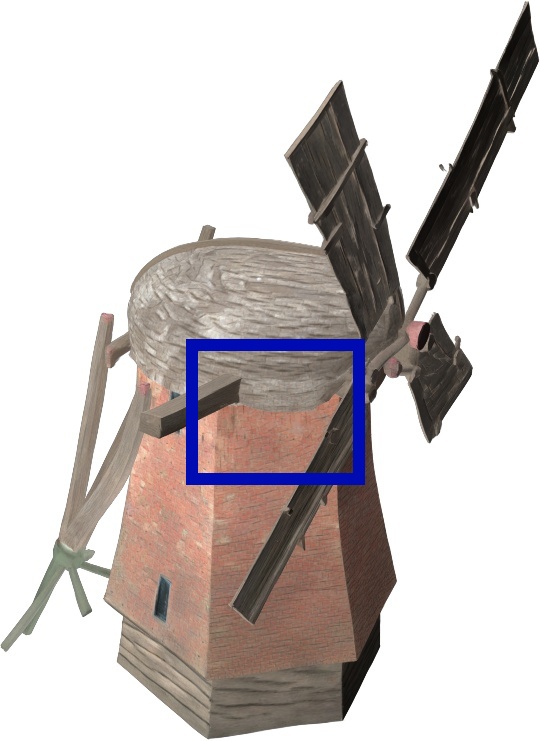} & \includegraphics[width=0.123\linewidth]{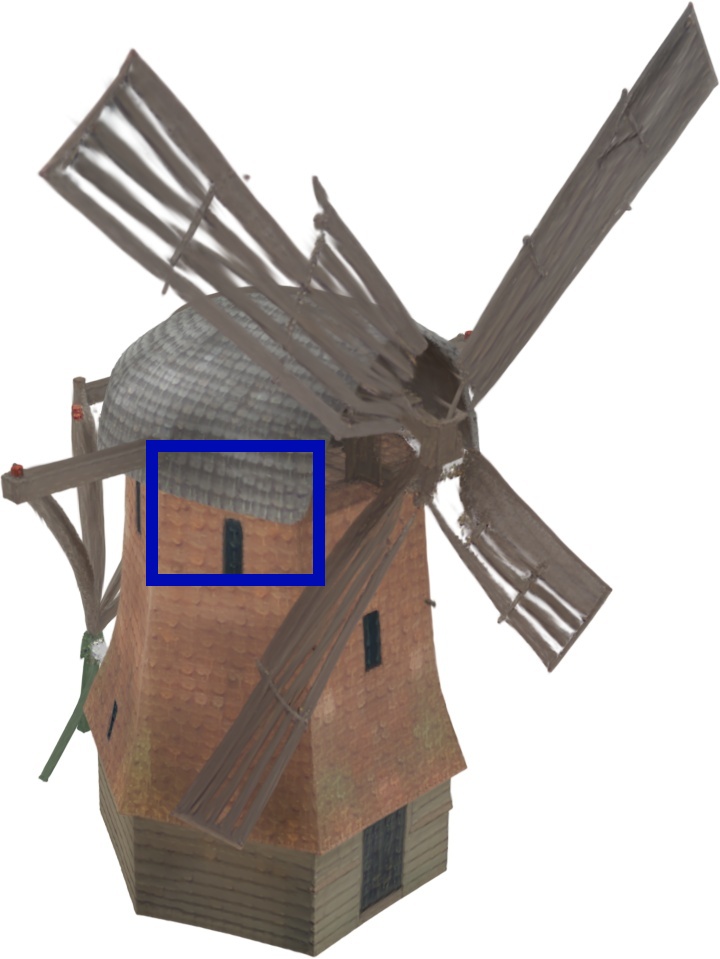} & \includegraphics[width=0.123\linewidth]{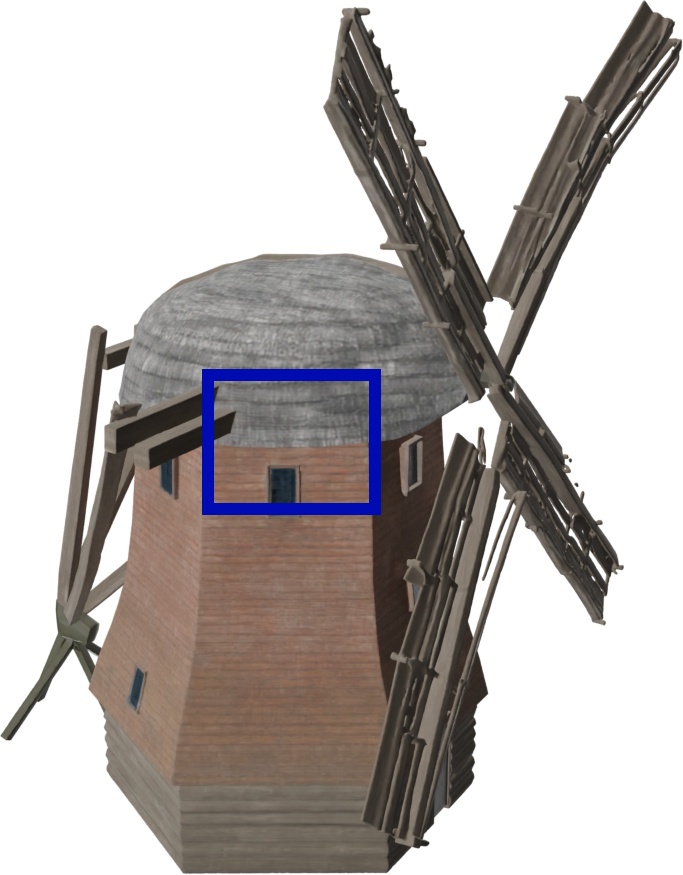}\\
            \includegraphics[width=0.123\linewidth]{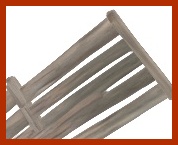} & \includegraphics[width=0.123\linewidth]{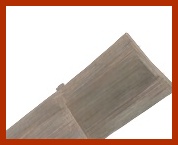} & \includegraphics[width=0.123\linewidth]{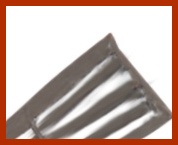} & \includegraphics[width=0.123\linewidth]{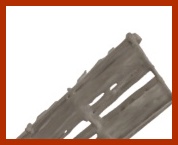} & \includegraphics[width=0.123\linewidth]{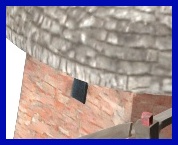} & \includegraphics[width=0.123\linewidth]{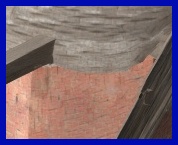} & \includegraphics[width=0.123\linewidth]{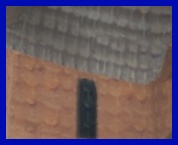} & \includegraphics[width=0.123\linewidth]{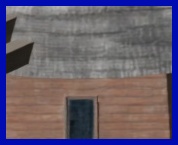}\\
            \footnotesize \textit{Split-and-scale} & \footnotesize LiTo-8192 & \footnotesize TRELLIS & \footnotesize TRELLIS-2 & \footnotesize \textit{Split-and-scale} & \footnotesize LiTo-8192 & \footnotesize TRELLIS & \footnotesize TRELLIS-2
        \end{tabular} \\
    \end{tabular}
    \caption{\textbf{Qualitative generation comparison.} A generative model trained on split-and-scale can reproduce fine details. In contrast, competitor methods suffer from either poor detail or poor material estimation. Asset credits~\cite{dessert,tree,windmill}.}
    \label{figure:generative_qual}
\end{figure}

\subsection{Generation}
\label{sec:gen}
\paragraph{Dataset.}\noindent We use the same dataset as reconstruction, but we render additional conditioning views at random distances away from the object, with field-of-view set such that the object takes up the majority of the frame.
We use two sets of cameras for evaluation. 
To capture the global structure of generated objects, we use ``horizon'' cameras sampled from a circle centered around the object, with a fixed field-of-view that captures the entire object silhouette. 
To measure the quality of finer details, we also use random poses that are randomly placed closer to the object than the horizon views. 
Our quantitative metrics are computed over the 2000 test set objects with highest aesthetic scores.

\begin{figure}[hbt]
    \centering
     \setlength{\tabcolsep}{0.5pt}
    \begin{tabular}{ccc@{\hskip 0.5in}cc}
        & \small estimated & \small ground truth\\
        \includegraphics[height=0.13\linewidth]{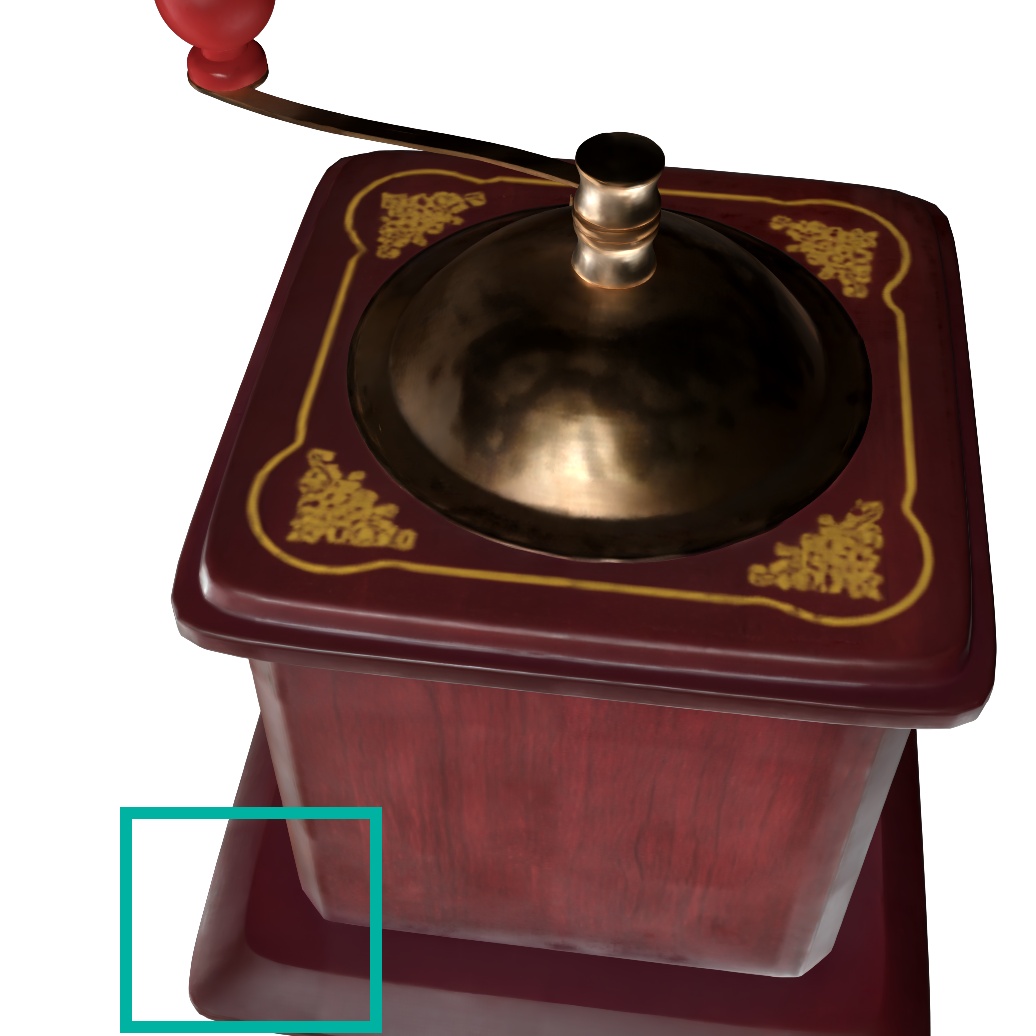} & \includegraphics[height=0.13\linewidth]{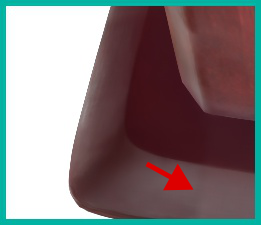} & \includegraphics[height=0.13\linewidth]{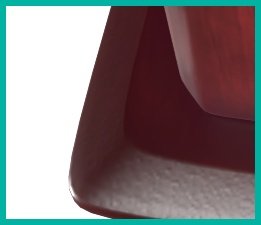} & 
        \includegraphics[height=0.13\linewidth]{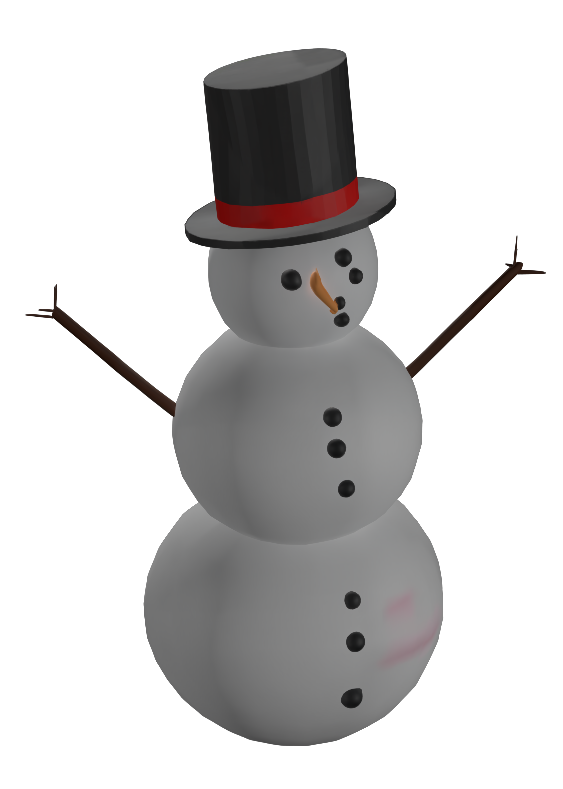} & \includegraphics[height=0.13\linewidth]{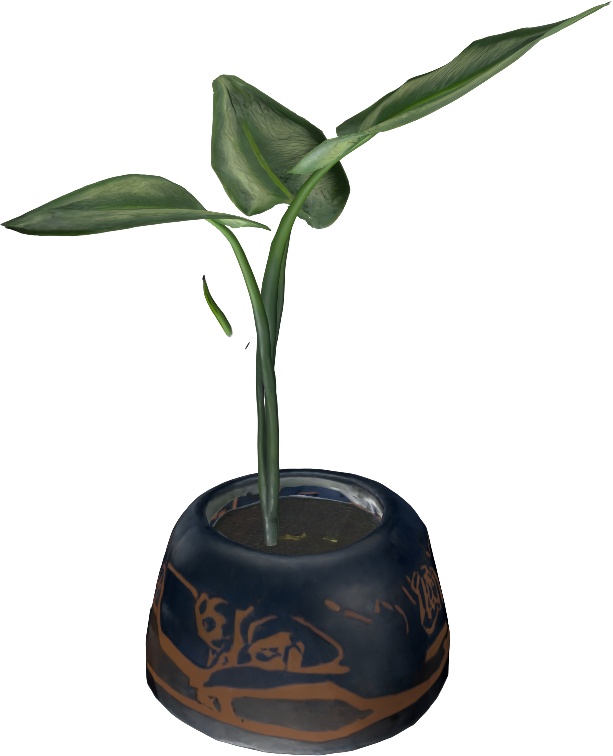} \\
        \multicolumn{3}{c}{\small {(a) Reconstruction seams}} & \multicolumn{2}{c}{\small {(b) Autoregressive structural flaws}}
    \end{tabular}
    \caption{\textbf{Failure cases.} \textbf{(a)} In tokenization, because each voxel is independently processed, subtle seams can appear in areas with significant view-dependent appearance. \textbf{(b)} Because our proposed generation model is autoregressive, sometimes it can experience the structural flaws common to such models. This manifests in the form of repeated geometry or poor symmetry. Asset credits~\cite{grinder,snowman,plant}.}
    \label{figure:failure_qual}
\end{figure}

\paragraph{Results.}\noindent \Cref{figure:generative_qual} shows example results from our proposed autoregressive  generation.  
More results can be found in the supplemental website. 
In general, we observe that our proposed method generates output assets with finer details than a 8192-token LiTo model (``LiTo-8192''), thanks to the increased capacity of the latent space. 
We also compare to generations from TRELLIS~\cite{xiang2025trellis} and TRELLIS-2~\cite{xiang2025trellis2}\footnote{We render TRELLIS-2’s textured mesh outputs with ground truth lighting (Section A.3).} --- our method seems to produce much less color shift relative to the input images compared to both methods. 
However, as shown in \Cref{figure:failure_qual}(b), we sometimes observe structural flaws in the output assets from our generative model, \eg, duplicated features or asymmetry. 
We believe such degradations are perhaps expected due to our autoregressive formulation, where later voxel generations cannot revise earlier ones.

For quantitative evaluation, \Cref{table:generation_table} shows the performance of our proposed autoregressive voxel generation method compared to existing 3D generation approaches. 
Our model can be seen to outperform TRELLIS and TRELLIS-2 baselines, while being competitive with LiTo-8192. 
In particular, our method seems to be roughly on par with LiTo-8192 in terms of input view metrics, while it begins to outperform for novel views. 

\begin{table}[ht]
    \centering
     \vspace{0.25em}
    \caption{\textbf{Quantitative generation comparison.} CFG scale is 3.}
    \label{table:generation_table}
    \adjustbox{max width=0.9\columnwidth}{%
    \begin{tabular}{c}
        \begin{tabular}{lcccccc}
            \toprule
            & \multicolumn{2}{c}{Input view} & \multicolumn{2}{c}{Novel view (horizon)} & \multicolumn{2}{c}{Novel view (random)} \\
            \cmidrule(lr){2-3} \cmidrule(lr){4-5} \cmidrule(lr){6-7}
            \textbf{Method} & \textbf{FID}$\downarrow$ & \textbf{KID$_{10^{-3}}$}$\downarrow$ & \textbf{FID}$\downarrow$ & \textbf{KID$_{10^{-3}}$}$\downarrow$ & \textbf{FID}$\downarrow$ & \textbf{KID$_{10^{-3}}$}$\downarrow$ \\

            LiTo-8192 & \first 16.78 & \second 0.42 & \second 6.33 & \second 0.97 & \second 7.95 & \first 1.77\\

            TRELLIS & {38.09} & {3.94} & {10.32} & {2.17} & {15.35} & {5.02} \\
            TRELLIS-2 & {42.84} & {3.76}&  {10.39} & {1.81} & 14.71 & {3.79} \\
            \textit{Split-and-scale} & \second 17.98 & \first 0.23 & \first 6.26 & \first 0.83 & \first 7.70 & \second 1.83\\
            \bottomrule
        \end{tabular} \\
    \end{tabular}%
    }
\end{table}

\section{Limitations and future work}
\label{sec:limits}
While split-and-scale is a promising approach for scalable latent representations, it is not without its limitations. 
First, because every subregion is encoded and decoded independently, subtle seam artifacts can arise, particularly in areas with significant view dependence (\Cref{figure:failure_qual}(a)). Visually, these artifacts appear as highlights that are subtly cut off at voxel boundaries. 
They can be mitigated in a compute-efficient way by training a decoder that conditions on neighboring latents. 
Second, the current implementation of split-and-scale subdivides objects into voxels of equal size for simplicity, regardless of the content complexity.
While our proposed adaptive latent allocation (see \Cref{sec:splitscaleimplementation}) can help mitigate any information asymmetry across different voxels, this regular division fundamentally increases the complexity of the distribution the split-and-scale tokenizer must be able to encode. 
Intelligent divisions, such as K-d tree or part-based approaches, could improve performance, but may introduce extra complexity for generative modeling or limit available training data.
Third, in our current split-and-scale implementation, the number of output Gaussians grows with the subdivision level (\ie, each decoder run produces a full-sized set of splats). For a large $n$, this can generate too many Gaussians to render efficiently in practice. A smarter implementation where the decoder can output a variable number of splats depending on $n$ would mitigate this problem.

Our training recipe for autoregressive generation is also quite simplistic, and can be further improved. 
For example, a common challenge with autoregressive models is that later generations cannot revise earlier ones, occasionally leading to global structural flaws (\Cref{figure:failure_qual}(b)). 
One scenario that occasionally appears is where the object looks reasonable from the input view, but occluded regions show slight asymmetries or floating geometry relative to the input.
A more complex design that enables the autoregressive model to generate global structures early in the sampling sequence could help remedy such issues, \eg, following recent work in fractal~\cite{li2025fractal} and discrete diffusion models. 
Another common challenge with such models is the requisite teacher forcing during training. Adapting recent techniques from other domains that attempt to deal with this training-inference gap~\cite{huang2025self,chen2024diffusion} could improve quality.

\section*{Acknowledgments}
We thank Byeongjoo Ahn, Raviteja Vemulapalli and Anagh Malik for their detailed feedback. All funding for this work came from Apple.

\section*{Author contributions}

All authors were involved in project planning and progress discussions.

\paragraph{Scope.} Rick Chang and Oncel Tuzel set the overarching direction of the work.

\paragraph{Method.} Rick Chang formulated the main split-and-scale methodology, and Dorian Chan formalized the computational benefits. Dorian Chan conceived the autoregressive generation architecture. Muhammed Kocabas, Xiaoming Zhao and Oncel Tuzel provided idea, design and architecture feedback.

\paragraph{Writing.} Dorian Chan wrote the initial draft of the paper, which Rick Chang and Muhammed Kocabas later refined. Oncel Tuzel critiqued the manuscript.

\paragraph{Model training.} Dorian Chan led model training for both the tokenizer and the generative model. Rick Chang and Xiaoming Zhao developed the original training framework.

\paragraph{Experiments.} Dorian Chan developed the evaluation framework for all ablations. Muhammed Kocabas implemented the comparisons with TRELLIS and TRELLIS-2. Dorian Chan and Rick Chang created all qualitative figures. Dorian Chan created the website.

\bibliographystyle{plainnat}
\bibliography{neurips/main}

\appendix
\section{Appendix}

\begin{table}[t]
    \centering
    \caption{\textbf{Ablation study.} We ablate our proposed designs for reconstruction as well as generation. In (b), both LiTo-32768 variants cannot be trained on an 80GB GPU, for both the representation and generative model. }
    \label{table:ablation_table}

    \begin{subtable}{\linewidth}
        \centering
        \adjustbox{max width=0.9\columnwidth}{%
        \begin{tabular}{lcccc}
            \toprule
              & \multicolumn{2}{c}{$n = 4$, $\approx 32768$ tokens} & \multicolumn{2}{c}{$n = 8$, $\approx 131072$ tokens}\\
              \cmidrule(lr){2-3} \cmidrule(lr){4-5}
              \textbf{{Recipe}} & \textbf{LPIPS} & \textbf{Chamfer} & \textbf{LPIPS} & \textbf{Chamfer}\\

            \midrule

            Full recipe & \first 0.107 & \first 2.19$\times 10^{-3}$ & \first 0.081 & \first 2.05$\times 10^{-3}$\\
            
            \quad no random point count & \second{} 0.110 & 2.23$\times 10^{-3}$ & 0.092 & 2.07$\times 10^{-3}$\\
            
            \quad no scaling & 0.161 & 2.92$\times 10^{-3}$ & 0.164 & 2.84$\times 10^{-3}$\\
            
            \midrule

            \multicolumn{5}{l}{\textbf{Training scales:}}\\
            \midrule

            Only $n=1$ and $n=2$ training & 0.113 & \second{} 2.21$\times 10^{-3}$ & \second{} 0.088 & \second{} 2.06$\times 10^{-3}$\\

            Only $n=1$ training & 0.145 & 3.34$\times 10^{-3}$ & 0.133 & 2.71$\times 10^{-3}$\\

            Only $n=4$ training & 0.135 & 2.29$\times 10^{-3}$ & 0.101 & 2.08$\times 10^{-3}$\\
            
            \bottomrule
        \end{tabular}%
        }
        \caption{Reconstruction ablation (44k iterations)}
        \label{table:ablation_recon}
    \end{subtable}

    \vspace{1ex}

    \begin{subtable}{\linewidth}
        \centering
        \def\maskedcolor{gray}
        \adjustbox{max width=0.9\columnwidth}{%
        \begin{tabular}{lcccccc}
            \toprule
            & \multicolumn{2}{c}{Input view} & \multicolumn{2}{c}{Novel view (horizon)} & \multicolumn{2}{c}{Novel view (random)} \\
            \cmidrule(lr){2-3} \cmidrule(lr){4-5} \cmidrule(lr){6-7}\\
             & \textbf{FID}$\downarrow$ & \textbf{KID}$\downarrow$ & \textbf{FID}$\downarrow$ & \textbf{KID}$\downarrow$ & \textbf{FID}$\downarrow$ & \textbf{KID}$\downarrow$
            \\
            \midrule
            \multicolumn{5}{l}{\textbf{Architecture (135k iterations):}}\\
            \midrule
            Autoregressive & \first{}21.56 & \first{}6.34$\times10^{-4}$ & \first{}8.63 &\first{}1.81$\times10^{-3}$ & \first{}11.67 & \first{}3.80$\times10^{-3}$\\
            Direct with smaller DiT & 27.67 & 7.35$\times10^{-4}$ & 21.89 & 8.65$\times10^{-3}$ & 32.63 & 1.58$\times10^{-2}$\\
            \textcolor{\maskedcolor}{LiTo-32768 DiT, equal time} & \textcolor{\maskedcolor}{21.95} & \textcolor{\maskedcolor}{7.35$\times10^{-4}$} & \textcolor{\maskedcolor}{8.76} & \textcolor{\maskedcolor}{1.62$\times10^{-3}$} & \textcolor{\maskedcolor}{11.38} & \textcolor{\maskedcolor}{3.44$\times10^{-3}$}\\ 
            \textcolor{\maskedcolor}{LiTo-32768 DiT, equal iter} & \textcolor{\maskedcolor}{18.50} & \textcolor{\maskedcolor}{2.30$\times10^{-4}$} & \textcolor{\maskedcolor}{6.53} & \textcolor{\maskedcolor}{8.29$\times10^{-4}$} & \textcolor{\maskedcolor}{7.95} & \textcolor{\maskedcolor}{1.77$\times10^{-3}$}\\ 
            \midrule
            \multicolumn{5}{l}{\textbf{Generation order:}}\\
            \midrule
            Random & 18.07 & 2.70$\times10^{-4}$ & \first 6.18 & 8.26$\times10^{-4}$ & 7.89 & 1.94$\times10^{-3}$\\
            Raster & \first 17.87 & \first 2.45$\times10^{-4}$ & 6.21 & \first 8.21$\times10^{-4}$ & \first 7.72 & \first 1.81$\times10^{-3}$\\
            \midrule
            \multicolumn{5}{l}{\textbf{Occupancy impact:}}\\
            \midrule
            GT occupancy & 17.95 & 2.48$\times10^{-4}$ & 6.15 & 7.78$\times10^{-4}$ & 7.75 & 1.83$\times10^{-3}$\\
            Generated occupancy & 17.98 & 2.26$\times10^{-4}$ & 6.26 & 8.29$\times10^{-4}$ & 7.70 & 1.83$\times10^{-3}$\\
            \bottomrule
        \end{tabular}%
        }
        \caption{Generation ablation}
        \label{table:ablation_gen}
    \end{subtable}
\end{table}

\begin{figure}[h]
    \centering
    \begin{tabular}{c}
        \includegraphics[width=\linewidth]{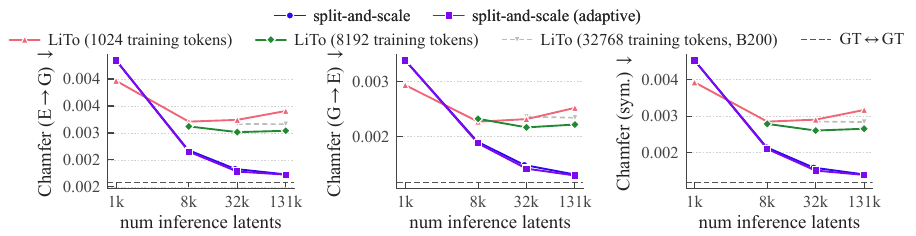} \\
        \textbf{(a) Equal compute time} \\\\
        \includegraphics[width=\linewidth]{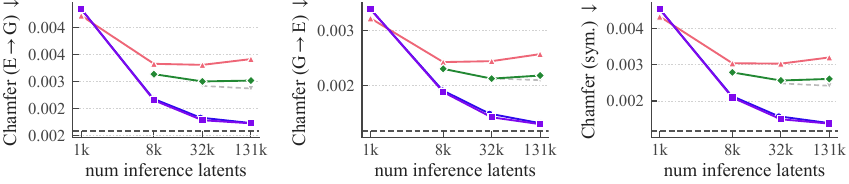} \\
        \textbf{(b) Equal iteration}
    \end{tabular}
    \caption{\textbf{Quantitative geometry comparison versus naive token count scaling.} We compare the quality of sampled point clouds from split-and-scale versus comparable LiTo models~\cite{lito}.}
    \label{figure:geometry_curves}
\end{figure}

\subsection{Additional experiments}

\Cref{table:ablation_table} details our ablation studies for both our tokenizer architecture as well as our generative model, as already discussed in the main manuscript. In the reconstruction ablation, one interesting point is that $n=4$-only training underperforms training with only $n=1$ and $n=2$. We believe the reasons to be twofold. First, coarse-level training may improve the quality of fine details; in particular, since we split the scene and scale each voxel back to the full range ($[-1, 1]$) and encode each voxel separately, each rescaled voxel of $n=4$ will contain more smoothed geometry compared to $n=1$ and $n=2$. For example, a voxel in $n=1$ may contain a chair, but a voxel in $n=4$ might only contain the seat back (which has lower geometry frequency). This means the training data in $n=4$ is of lower frequency than $n=1$ or $n=2$. Thus, training with $n=1$ gives the highest content frequency, which is important for handling any downstream scenarios with extremely detailed objects. To handle boundaries between voxels, it is important to train with $n=2$ and $n=4$, as done in the full recipe. A second potential reason is compute-related implementation tradeoffs: specifically, in order to fit training within 80GB GPU memory (e.g., A100 or H100 GPU) as well as increase speed, we detach all but a single voxel at every iteration when $n=4$ (see \Cref{sec:splitscaleimplementation}). In contrast, we can jointly train 8 voxels at a time when $n=2$, resulting in a stronger training signal. This difference in trainable voxels across different $n$ can be attributed to the larger number of Gaussian activations that must be stored for backprop when more voxels are used, even when they are detached. This effect is validated in a small-scale experiment in \Cref{table:n_ablation_small}.

\begin{figure}[t]
    \centering
    \begin{tabular}{c}
        \includegraphics[width=\linewidth]{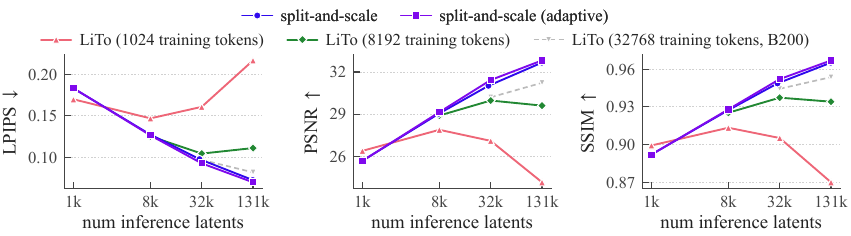} \\
        \textbf{(a) Equal compute time} \\\\
        \includegraphics[width=\linewidth]{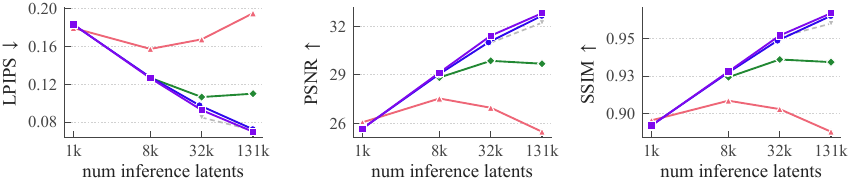} \\
        \textbf{(b) Equal iteration}
    \end{tabular}
    \caption{\textbf{Quantitative comparison versus naive token count scaling.} Following the recipe in \Cref{subsec:splitscale}, we train a split-and-scale model that outputs 1024 latents per voxel at inference. For comparison, we train 1024 and 8192 token LiTo models~\cite{lito} that use a similar amount of memory/compute during training. In both (a) equal compute time and (b) equal iteration scenarios, split-and-scale can be seen to scale with token count better than the baseline models. Although it cannot be trained within the memory limits of an 80GB GPU, we include a 32768 token variant of LiTo~\cite{lito} as a performance upper-bound; split-and-scale is still competitive.}
    \label{figure:tokenizer_curves_full}
\end{figure}

\Cref{figure:geometry_curves} and \Cref{figure:tokenizer_curves_full} are extensions to \Cref{figure:tokenizer_curves}.
\Cref{figure:tokenizer_curves_full} shows the results under an equal compute setting, where LiTo-1024 has been trained to 200k iterations, LiTo-8192 to 100k iterations, LiTo-32768 to 34k iterations and split-and-scale to the standard 90k iterations. Split-and-scale can be seen to be very strong. \Cref{figure:geometry_curves} explores the scaling performance of split-and-scale versus naive token scaling in the context of geometry only --- split-and-scale can be seen to significantly outperform the standard approach, in both equal compute and equal iteration settings. Both figures also show the impact of adaptive allocation, as will be discussed in \Cref{sec:splitscaleimplementation}. 

\begin{figure}[h]
    \centering
     \setlength{\tabcolsep}{0.5pt}
     \def\bgcolor{gray!80}
     \def\fgcolor{white}
     \begin{tabular}{ccc@{\hspace{2ex}}|@{\hspace{2ex}}ccc}
        \includegraphics[width=0.14\linewidth]{neurips/figures/tokenizer_qual/0ee4f66bc63e477bb81673d7e3e633c6-r000/gt_000_box.jpg} &  \includegraphics[width=0.14\linewidth]{neurips/figures/tokenizer_qual/0ee4f66bc63e477bb81673d7e3e633c6-r000/split_scale_scratch_g124_90k_1024_4_gt_tokenizer_adaptive32768_recon_000_box.jpg} \put(-40, 45){\colorbox{\bgcolor}{\color{\fgcolor} \tiny LPIPS=0.16}}
        & \includegraphics[width=0.14\linewidth]{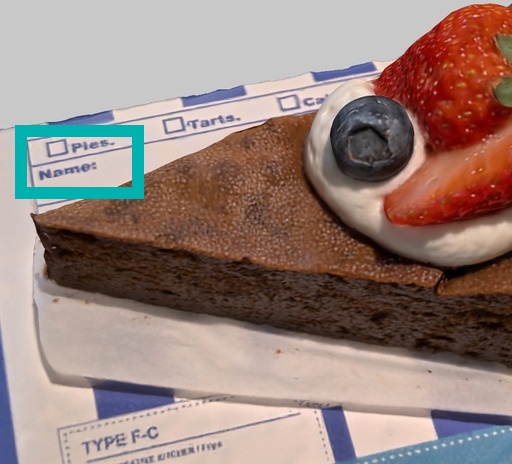} \put(-40, 45){\colorbox{\bgcolor}{\color{\fgcolor} \tiny LPIPS=0.14}} & \includegraphics[width=0.14\linewidth]{neurips/figures/tokenizer_qual/36b3656d60d44061926820c4ffdbbaf5-r000/gt_007_box.jpg} & \includegraphics[width=0.14\linewidth]{neurips/figures/tokenizer_qual/36b3656d60d44061926820c4ffdbbaf5-r000/split_scale_scratch_g124_90k_1024_4_gt_tokenizer_adaptive32768_recon_007_box.jpg} \put(-40, 45){\colorbox{\bgcolor}{\color{\fgcolor} \tiny LPIPS=0.08}}&
        \includegraphics[width=0.14\linewidth]{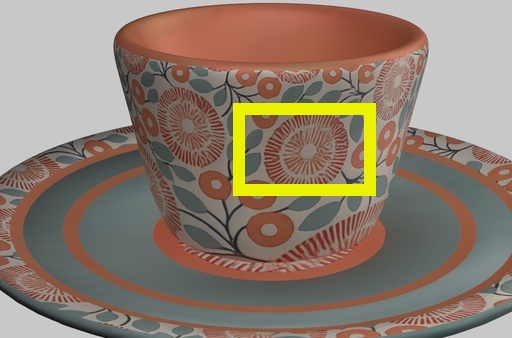} \put(-40, 45){\colorbox{\bgcolor}{\color{\fgcolor} \tiny LPIPS=0.06}} 
        \\[-1ex]
        \includegraphics[width=0.14\linewidth]{neurips/figures/tokenizer_qual/0ee4f66bc63e477bb81673d7e3e633c6-r000/gt_000_inset.jpg} &  \includegraphics[width=0.14\linewidth]{neurips/figures/tokenizer_qual/0ee4f66bc63e477bb81673d7e3e633c6-r000/split_scale_scratch_g124_90k_1024_4_gt_tokenizer_adaptive32768_recon_000_inset.jpg}  
        & \includegraphics[width=0.14\linewidth]{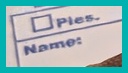} &   \includegraphics[width=0.14\linewidth]{neurips/figures/tokenizer_qual/36b3656d60d44061926820c4ffdbbaf5-r000/gt_007_inset.jpg} & \includegraphics[width=0.14\linewidth]{neurips/figures/tokenizer_qual/36b3656d60d44061926820c4ffdbbaf5-r000/split_scale_scratch_g124_90k_1024_4_gt_tokenizer_adaptive32768_recon_007_inset.jpg} &\includegraphics[width=0.14\linewidth]{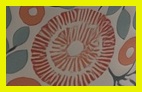} \\
        \footnotesize GT & 
        \begin{tabular}{c}\textit{\footnotesize Split-and-scale} \\\footnotesize (32768 inf.)\end{tabular} 
        & \begin{tabular}{c}\footnotesize \textcolor{gray!30}{LiTo-32768}\\\footnotesize \textcolor{gray!30}{(32768 inf.)}\end{tabular} & \footnotesize GT & 
        \begin{tabular}{c}\textit{\footnotesize Split-and-scale} \\\footnotesize (32768 inf.)\end{tabular} 
        & \begin{tabular}{c}\footnotesize \textcolor{gray!30}{LiTo-32768}\\\footnotesize \textcolor{gray!30}{(32768 inf.)}\end{tabular} \\[2ex]

    \end{tabular}

    \begin{tabular}{ccc}

        \includegraphics[width=0.14\linewidth]{neurips/figures/tokenizer_qual/0af4bfd183b24d578341157e1775ab1c-r000/gt_005_box.jpg} & \includegraphics[width=0.14\linewidth]{neurips/figures/tokenizer_qual/0af4bfd183b24d578341157e1775ab1c-r000/split_scale_scratch_g124_90k_1024_4_gt_tokenizer_adaptive32768_recon_005_box.jpg} \put(-40, 64){\colorbox{\bgcolor}{\color{\fgcolor} \tiny LPIPS=0.13}} &
        \includegraphics[width=0.14\linewidth]{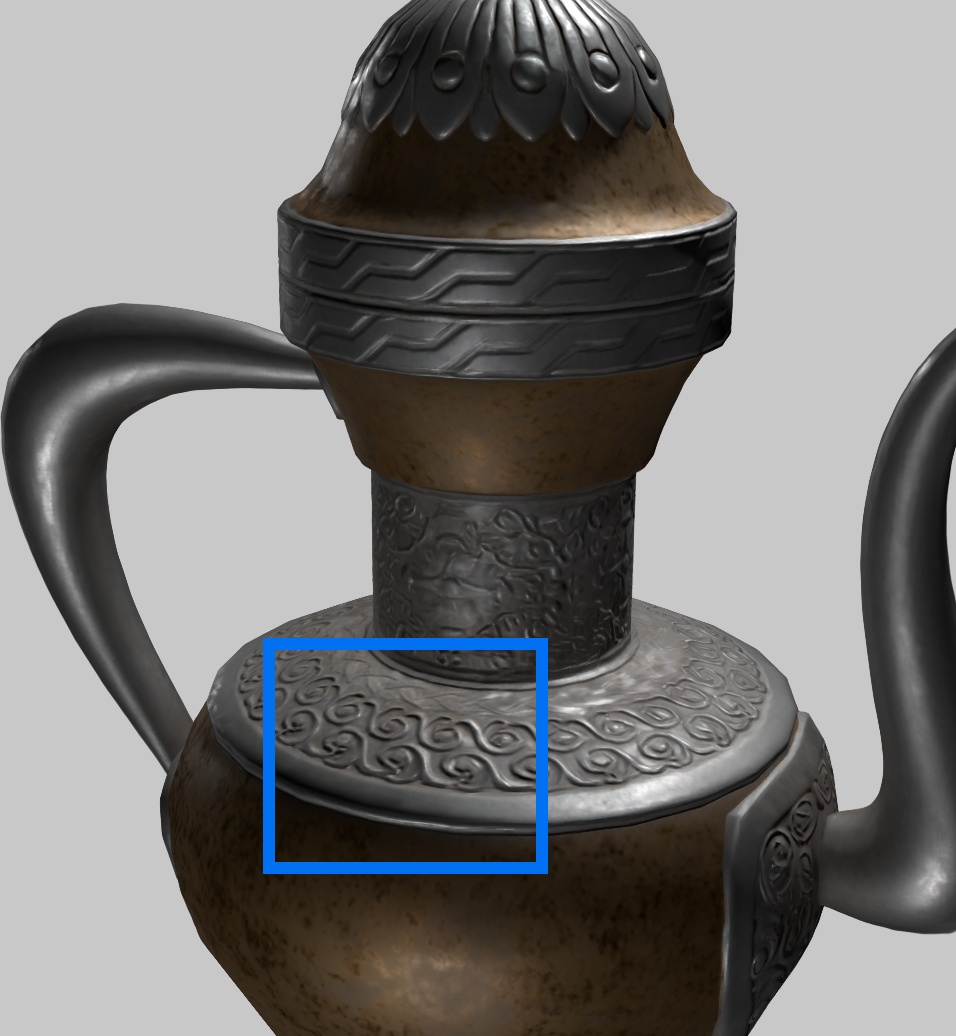} \put(-40, 64){\colorbox{\bgcolor}{\color{\fgcolor} \tiny LPIPS=0.10}}\\[-1ex]
        \includegraphics[width=0.14\linewidth]{neurips/figures/tokenizer_qual/0af4bfd183b24d578341157e1775ab1c-r000/gt_005_inset.jpg} & \includegraphics[width=0.14\linewidth]{neurips/figures/tokenizer_qual/0af4bfd183b24d578341157e1775ab1c-r000/split_scale_scratch_g124_90k_1024_4_gt_tokenizer_adaptive32768_recon_005_inset.jpg} & \includegraphics[width=0.14\linewidth]{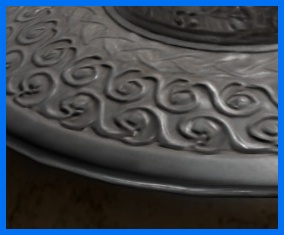} \\
        \footnotesize GT & 
        \begin{tabular}{c}\textit{\footnotesize Split-and-scale} \\\footnotesize (32768 inf.)\end{tabular} 
        & \begin{tabular}{c}\footnotesize \textcolor{gray!30}{LiTo-32768}\\\footnotesize \textcolor{gray!30}{(32768 inf.)}\end{tabular} 
    \end{tabular}
    \caption{\textbf{Upper-bound qualitative reconstruction comparison.} Building off \Cref{figure:tokenizer_qual}, although it cannot be trained within the memory limits of an 80GB GPU, we show a 32768 token variant of LiTo (``LiTo-32768'') as an upper-bound for performance; split-and-scale is still competitive. Asset credits~\cite{cake,teacup,jug}.}
    \label{figure:upper_tokenizer_qual}
\end{figure}

In \Cref{figure:upper_tokenizer_qual}, we also show an additional qualitative comparison of split-and-scale with 32768 inference tokens, and the upper-bound LiTo-32768 which cannot be trained in an 80GB GPU (\Cref{table:tokenizer_compute}). Split-and-scale remains competitive in visual performance.

To validate the importance of our multiscale training recipe, we also test whether the pretrained TRELLIS~\cite{xiang2025trellis} tokenizer can be used in a subdivision, tile-based fashion out of the box with $n=2$ and $n=4$. Specifically, as the base TRELLIS uses a $64^3$ grid, for $n=2$, we construct a $128^3$ grid and reconstruct individual $64^3$ subgrids using TRELLIS. While the level of fine detail can be seen to qualitatively improve, large blur artifacts appear along object and voxel boundaries that worsen with $n$. Introducing overlapping reconstructions across different voxels does not seem to mitigate these issues and in fact worsens performance --- blurry tile seams actually cover larger portions of the object than if they were not overlapped. This shows that tiling-aware training, as we developed in this paper, is important for scaling latent sizes as we have proposed.
\subsection{Split-and-scale implementation details}
\label{sec:splitscaleimplementation}

\paragraph{Latent representation recipe.}\noindent
Our work builds off the LiTo representation~\cite{lito}, with a Perceiver encoder, a Gaussian decoder, and a flow matching-based point-cloud decoder~\cite{chang20243d}. 
We randomly select a grid size of $n=1$, $2$ or $4$ at every iteration; we train for a total of 90k iterations. 
When $n=4$ ($\approx$ 32 active voxels on average), we detach all but one voxel. 
During training, we select anywhere between $2^{10}$-$2^{20}$ total input points, and randomly select a latent size between 256-4096 tokens per voxel. 
For supervision, we decode Gaussians for every voxel, render them jointly, and compute PSNR and LPIPS losses against the ground truth. We also apply a flow-matching loss through the point-cloud decoder~\cite{chang20243d} for every voxel.
During inference we use 1024 tokens per voxel, and a maximum of 1 million points per voxel.

One potential downside of naive split-and-scale is that a fixed $m$ latent tokens are used per voxel, when some voxels may contain more details and content than others. To tackle this, we can also train the encoder to handle a variable number of latent tokens, and at inference select the number of tokens per voxel according to some metric, \eg, the number of input points per voxel. The improvement afforded by this adaptive allocation is shown in \Cref{figure:tokenizer_curves_full}. Unless otherwise noted, we use this variant for all reconstruction results shown in the paper. For generation, we do not use this adaptation to avoid any additional prerequisite generation outside coarse $4\times4\times4$ occupancy. 

\subsubsection{Network architectures}

Our work is built off LiTo~\cite{lito}, and primarily inherits the same encoder/decoder/generative architectures from their work. Please refer to their paper for detailed architecture diagrams. Here, we summarize how each one works, and any modifications we have made.

\paragraph{Encoder:}\noindent A Perceiver encoder receives a surface light field as input, which it subsamples to use as queries. Each query cross attends to only the points that it is the closest query to --- this configuration can be interpreted as a sort of 3D patchification. Instead of the voxel-based attention pattern leveraged by LiTo for computational efficiency, we instead use standard full attention, as our per-voxel latent sizes are small enough for it to not become a bottleneck. 

\paragraph{Point-cloud decoder:}\noindent A noisy input point is positionally encoded and then passed into a Transformer architecture, where it only cross attends to the corresponding voxel latent (there is no self attention). The model conditions on the flow-matching timestep via gating and layernorm parameters predicted by a small MLP.

\paragraph{Gaussian decoder:}\noindent A Perceiver encoder receives $64^3$ coarse occupancy as input, cross attending to the latent of the corresponding voxel. It decodes a feature for every coarse location, each of which is then processed by an MLP to produce 64 Gaussians. Note that this decoder runs once for every occupied voxel, \ie, the number of Gaussians scales with the number of occupied voxels and therefore latent capacity.

\paragraph{Voxel decoder:}\noindent A Perceiver architecture is used to decode a sparse structure TRELLIS latent. A pretrained TRELLIS sparse-structure decoder is then used to convert this latent to a full occupancy grid. To train this voxel decoder, at every iteration we select a random voxel, compute a split-and-scale latent for it and then decode a single sparse structure latent --- we do not predict sparse structure latents for all voxels as we found such a procedure to be too memory inefficient for 80GB GPU training. We use this voxel decoder to decode Gaussians from generated latents.

\paragraph{Mesh decoder:}\noindent A Perceiver takes in coarse occupancy for queries (like the Gaussian decoder) and outputs low resolution shape features, which are then upsampled by sparse convolutional layers to Flexicube parameters~\cite{he2025triposf,shen2023flexible}. We follow the split-and-scale recipe described in the main paper to train this mesh decoder, \ie, we pick a random $n$, decode up to $n^3$ sets of Flexicube parameters, construct a mesh from their union, and compute losses on rendered depth and silhouettes. We do not compute losses on rendered normal, because we observe that the normals and depths of synthetic geometry do not often align (\eg, bump or normal maps). 

\paragraph{Generative model:}\noindent Like LiTo, our work leverages a standard Diffusion Transformer architecture, but to sample in an autoregressive fashion. Unlike LiTo, we do not use a positional embedding for each noisy token, but instead a ``voxel'' embedding on which voxel each token should correspond to --- we use a sinusoidal positional embedding of the voxel center, with 32 frequencies sampled from $2^0$ to $2^{12}$. We use a hidden dimension of 1152, with 16 self attention and cross attention heads. Time conditioning is implemented via adaLN-Zero~\cite{peebles2023scalable}. We input in $\mathbf{v}_\text{context}$ and $I$ via cross attention, where the image is preprocessed with DINOv2-large~\cite{oquab2023dinov2} and a learnable patchification layer, following LiTo~\cite{lito}. 
$I$ and $\mathbf{v}_\text{context}$ are given separate key-value and output projections, and their output cross attentions are summed.
$O$ is input via self attention with a single positionally-encoded learnable token per active voxel.
To generate $O$, we use the same flow-matching architecture as above, with $O$ represented as $4^3$ binary tokens --- we threshold with 0.5 to obtain the active voxels. We train this occupancy generator for 100k iterations.

\subsection{Baseline implementation Details}
\label{sec:implementation}

\paragraph{LiTo~\cite{lito}:}\noindent We retrain LiTo variants with the same training data as split-and-scale. We primarily reuse the exact Perceiver-IO architectures used by LiTo for the encoder and decoder --- we also use the same architectures for our split-and-scale encoder/decoder for fair comparison. When training variants with different latent counts, we also adjust the resolution of the occupancy grid that is input to the Gaussian decoder, in order to match the equivalent occupancy grid resolution used by split-and-scale for a particular latent count, \ie, 1024 tokens $\rightarrow$ $64^3$, 8192 tokens $\rightarrow$ $128^3$, 32768 tokens $\rightarrow$ $256^3$. We use ground truth occupancy as input to the Gaussian decoder for reconstruction results, and the output of a voxel decoder for generation results.

For the generative model, we use the same DiT architecture as LiTo~\cite{lito}, except without positional embedding to aid with set generation. We also use $1036\times1036$ image inputs, rather than the $518\times518$ used in the original paper for fairness. We train this model for a total of $240$k iterations --- note that our AR model, at $450$k iterations, has only been trained for roughly half the total object data of this LiTo model, as it is trained to generate only a subset of a given sample at every iteration.

\paragraph{TRELLIS~\cite{xiang2025trellis}:}\noindent
We use the publicly released TRELLIS pipelines for image-to-3D and the tokenizer. For the tokenizer,
TRELLIS expects a sparse $64^3$ occupancy grid plus per-view DINO features.
We voxelize the ground-truth point cloud directly into TRELLIS's $64^3$ format. The DINO pool consumes 150 dense input views (\ie, the same views used to generate the point cloud). We evaluate the results using TRELLIS' own Gaussian decoder. We render the output gaussians at our target cameras using TRELLIS's own splat rasterizer for a fair comparison against split-and-scale renders.

For image-to-3D, TRELLIS produces results in its own canonical frame ($[-0.5, 0.5]^3$, y-up), which does not align with our input-view coordinate system; we account for this with a fixed similarity transform (axis flip and $\times 2$ scale) before rendering. This rigid alignment is exact for the tokenizer (which is conditioned on the GT geometry), but for the generative model the orientation of the asset relative to the camera is not constrained by the conditioning image. We therefore only report quantitative gen-numbers using set-level metrics (FID/KID) that are invariant to such a global rotation. For qualitative results, we manually realign each gen output.

\paragraph{TRELLIS-2~\cite{xiang2025trellis2}:}\noindent
We use the publicly released checkpoints of TRELLIS-2. The tokenizer has the spatial downsample
factor of $16$ and $32$ latent channels per active voxel. We run the
tokenizer at $1024^3$ voxel resolution, which after the
$f16$ downsample yields a $64^3$ sparse latent grid populated only at
voxels intersected by the surface.

For image-to-3D, we run TRELLIS-2 with the $1024^3$ voxel resolution.
Because TRELLIS-2 represents appearance through PBR materials, evaluating
it against our Gaussian renderings requires re-rendering all
TRELLIS-2 outputs under the same lighting as our GT. We therefore re-render
every reconstructed/generated GLB through Blender Cycles, using the exact
per-sample HDRI light that were used to render our GT views. In practice TRELLIS-2's
material-light disentanglement is not perfect: even with the GT lighting,
the decoded base-color/metallic/roughness channels often produce
appearance noticeably different from our diffuse GT, which inflates
TRELLIS-2's pixel-space and FID/KID numbers in ways that are not entirely
attributable to the tokenizer/generator itself. 

As with TRELLIS,
TRELLIS-2's gen output lives in its own canonical frame; we apply a fixed
rotation/scale alignment for the qualitative figures, and rely on
rotation-invariant set metrics (FID/KID) for the quantitative gen
comparison. All renders use $1036{\times}1036$ resolution to match
split-and-scale's GT.

\begin{table}[htbp]
    \centering
    \caption{\textbf{Few-iteration training scale ablation.} We evaluate the impact of training at $n=4$ without detaching any voxels. While this makes training infeasible on most standard 80GB GPUs, it provides the best performance. }
    \label{table:n_ablation_small}
    \adjustbox{max width=0.9\columnwidth}{%
    \begin{tabular}{lccccc}
        \toprule
        \textbf{Training recipe} & \textbf{Fits 80 GB?} & $n=4$ \textbf{LPIPS} & $n=4$ \textbf{PSNR} & $n=8$ \textbf{LPIPS} & $n=8$ \textbf{PSNR}\\
        \midrule
        Our recipe & yes & 0.129 & 29.58 & 0.102 & 31.26\\
        Only $n=1$ training & yes & 0.168 & 27.60 & 0.158 & 28.82\\
        Only $n=1$ and $n=2$ training & yes & 0.131 & 29.61 & 0.106 & 31.37\\
        Only $n=4$ training, detach & yes & 0.161 & 27.17 & 0.128 & 28.98\\
        Only $n=4$ training, no detach & no & 0.111 & 29.72 & 0.080 & 31.69\\
        \bottomrule
    \end{tabular}
    }
\end{table}

\begin{table}[htbp]
    \centering
    \caption{\textbf{TRELLIS tiling experiment.} We evaluate whether the vanilla TRELLIS~\cite{xiang2025trellis} tokenizer can be applied in a similar tile-based fashion as split-and-scale. Quantitatively, performance can be seen to drop with the amount of tiling, even with overlapping reconstructions across different voxels. }
    \label{table:trellis_tiling}
    \begin{tabular}{lcc}
        \toprule
        \textbf{Method} & \textbf{LPIPS} & \textbf{PSNR}\\
        \midrule
        original TRELLIS~\cite{xiang2025trellis} & 0.212 & 20.86\\
        TRELLIS, $n=2$ & 0.254 & 19.63\\
        TRELLIS, $n=2$ + overlapping & 0.284 & 18.69\\
        TRELLIS, $n=4$ & 0.295 & 17.86\\
        \bottomrule
    \end{tabular}
\end{table}

\newpage

\end{document}